\documentclass[11pt]{article}

\usepackage[preprint]{acl}

\usepackage{times}
\usepackage{latexsym}
\usepackage{algorithmicx,algorithm}

\usepackage{tikz}
\usetikzlibrary{shapes, arrows.meta, positioning, calc, fit, backgrounds}
\usepackage{fontawesome5} % 用于图标，如果没有安装可以去掉或换成文字

\usepackage[utf8]{inputenc}

\usepackage{CJKutf8}

\usepackage{microtype}

\usepackage{url}
\usepackage{inconsolata}

\usepackage{graphicx}

\usepackage{amsmath}
\usepackage{amssymb}

\usepackage{booktabs}
\usepackage{multirow}

\usepackage{tcolorbox}
\tcbuselibrary{breakable,skins}

\usepackage{xcolor}
\usepackage{colortbl}

\usepackage{subcaption}

\title{Rethinking Multiple-Choice Questions for RLVR: Unlocking Potential via Distractor Design}

\author{
  Xu Guo$^{*1,2,3}$ \quad
  Qiming Ge$^{*1,3}$ \quad
  Jian Tong$^{1}$ \quad
  Kedi Chen$^{1,2}$ \quad
  Jin Zhang$^{1,3}$ \quad
  Xiaogui Yang$^{1}$ \\
  Xuan Gao$^{1,3}$ \quad
  Haijun Lv$^{1}$ \quad
  Zhihui Lu$^{\dagger 3}$ \quad
  Yicheng Zou$^{1}$ \quad
  Qipeng Guo$^{\dagger 1}$ \\[6pt]
  $^{1}$Shanghai AI Laboratory \quad
  $^{2}$Shanghai Innovation Institute \quad
  $^{3}$Fudan University \\[4pt]
  {\small \texttt{\{geqiming, tongjian, chenkedi, zhangjin, yangxiaogui,}} \\
  {\small \texttt{gaoxuan, lvhaijun, zouyicheng, guoqipeng\}@pjlab.org.cn}} \\
  {\small \texttt{guox24@m.fudan.edu.cn} \quad \texttt{lzh@fudan.edu.cn}} \\[2pt]
  {\small $^*$Equal contribution \quad $^\dagger$Corresponding authors}
}

\begin{document}
\begin{CJK}{UTF8}{gbsn}  % 开始中文环境, 使用 gbsn 字体（宋体）

\maketitle

% \begin{abstract} 
% Reinforcement Learning with Verifiable Rewards (RLVR) benefits from precise reward signals. 
% Multiple-choice questions (MCQs) seem ideal, being broad and verifiable. 
% However, their effectiveness is undermined by random guessing and answer elimination, which corrupt reward signals. 
% Existing approaches simply discard MCQs or convert them to open-ended formats, discarding distractors (the incorrect options) as useful negative samples. 
% We instead leverage distractors as natural negative signals, revisiting MCQs for RLVR through the lens of option design. 
% We study two factors: the quantity of distractors and their strength. 
% From our analysis, two observations emerge: (1) on average, option count matters little; what matters is the train-test match; (2) distractor quality dominates—a strong distractor makes even 2-way questions effective for RLVR. 
% Guided by these findings, we propose a simple Iterative Distractor Curation (IDC) framework that enhances distractor quality while keeping the original question intact.
% Surprisingly, models can self-generate plausible distractors to improve their own training.
% Across SuperGPQA, MMLU-Pro, MedXpertQA, and PubMedQA, this method consistently improves performance, demonstrating that systematically curated distractors enhance RLVR training effectiveness.
% \end{abstract}

\begin{abstract} 
Reinforcement Learning with Verifiable Rewards (RLVR) significantly enhances the reasoning capabilities of Large Language Models. 
When applied to RLVR, Multiple-Choice Questions (MCQs) offer a scalable source of verifiable data but risk inducing reward hacking, where models shortcut reasoning via random guessing or simple elimination. 
Current approaches often mitigate this by converting MCQs to open-ended formats, thereby discarding the contrastive signal provided by expert-designed distractors. 
In this work, we systematically investigate the impact of option design on RLVR. 
Our analysis highlights two primary insights: (1) Mismatches in option counts between training and testing degrade performance. 
(2) Strong distractors effectively mitigate random guessing, enabling effective RLVR training even with 2-way questions. 
Motivated by these findings, we propose \textbf{Iterative Distractor Curation} (IDC), a framework that actively constructs high-quality distractors to block elimination shortcuts and promote deep reasoning. 
Experiments on various benchmarks demonstrate that our method effectively enhances distractor quality and yields significant gains in RLVR training compared to the original data. 
\end{abstract}

\section{Introduction}

Reinforcement Learning with Verifiable Rewards (RLVR) has recently improved large language models (LLM) reasoning capabilities~\citep{lambert2025tulu3pushingfrontiers,shao2024deepseekmathpushinglimitsmathematical,chen2025surveyinductivereasoninglarge}.
Unlike RLHF~\citep{ouyang2022traininglanguagemodelsfollow}, RLVR uses automatically verifiable rewards, eliminating human annotation. 
The performance of RLVR depends on the accuracy of the reward signal: an LLM samples diverse outputs, receives contrastive feedback, and updates its policy accordingly. 
When the task outputs admit clear correctness criteria (for example, when answers can be uniquely verified or 
logical reasoning can be checked via formal verification), RLVR can effectively improve model performance.

\begin{figure}[t]
    \centering
    \includegraphics[width=\columnwidth]{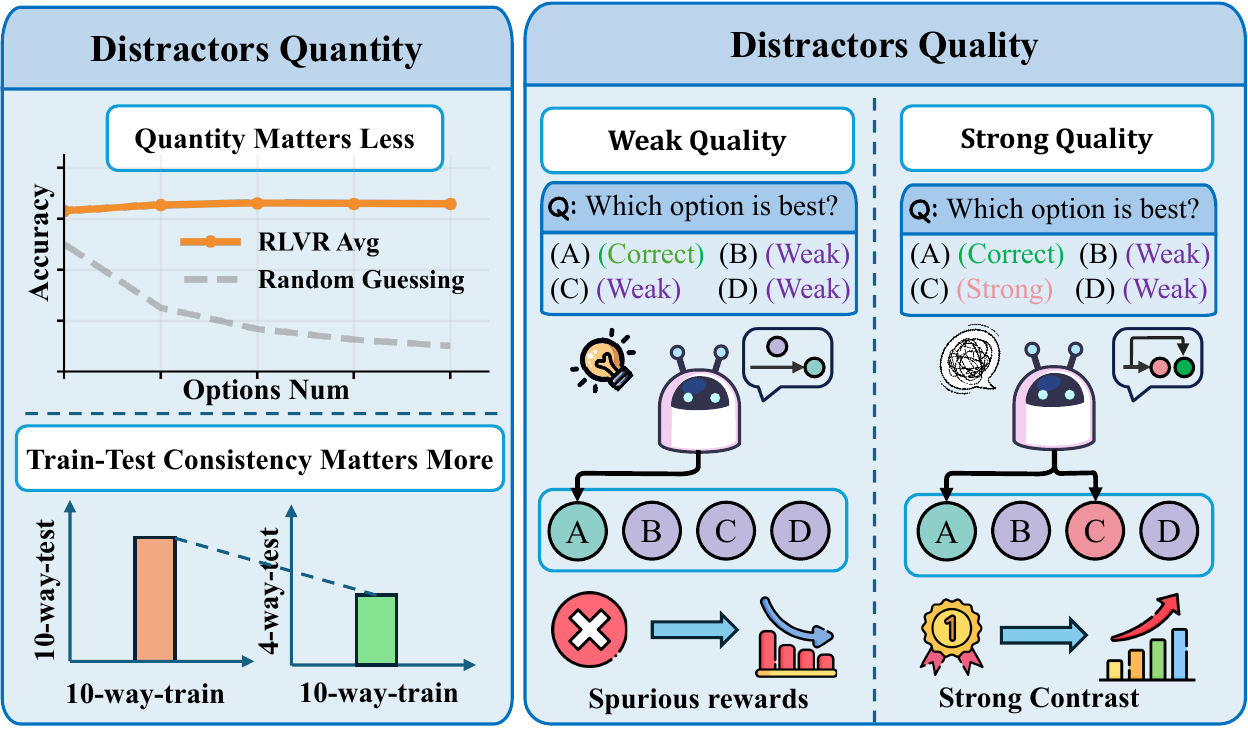}
    \caption{The impact of distractor design on RLVR.}
    \label{fig:teaser}
\end{figure}

However, not all tasks provide reliable verification. Recent work has explored Multiple-Choice Questions (MCQs) as verifiable reward signals to enable automated training across a broader range of tasks~\citep{liu2025distillationpushinglimitsmedical,zhang2025medrlvremergingmedicalreasoning,akter2025nemotroncrossthinkscalingselflearningmath,guo2025seed15vltechnicalreport}.
For example, AlphaMed~\citep{liu2025distillationpushinglimitsmedical} performs RLVR with medical MCQs and obtains performance gains.
Multiple-choice questions are typically considered verifiable, as their correctness can be objectively determined by comparing against the reference answer.
Nevertheless, Med-RLVR~\citep{zhang2025medrlvremergingmedicalreasoning} finds that models can hack rewards~\citep{amodei2016concreteproblemsaisafety} by shortcutting reasoning to output answers directly, highlighting the risks of using MCQs for verification.
The accuracy of reward signals in MCQ-based RLVR is inherently challenged by two factors: (1) \textbf{Random guessing}: due to limited answer space, models can guess the correct answer even for unsolved questions; and (2) \textbf{Elimination strategies}: models can exploit weak distractors (i.e., options that are obviously incorrect) to identify correct answers via elimination rather than reasoning.

To address these issues, Kimi k1.5~\cite{kimiteam2025kimik15scalingreinforcement} and Seed-Thinking-v1.5~\citep{guo2025seed15vltechnicalreport} either discard multiple-choice questions or convert them into open-ended formats. Similarly, Nemotron-CrossThink~\citep{akter2025nemotroncrossthinkscalingselflearningmath} selectively converts suitable MCQs into QA pairs, filtering out those that depend on the provided options (e.g., ``Which of the following...''). 
These studies attribute the limitations of MCQ-based RLVR to the closed-ended format: the presence of options increases the probability of sampling correct answers by chance.
To circumvent this, they transform MCQs into open-ended formats rather than training on the original questions.
However, this approach discards the benefits of MCQs: beyond the correct option, MCQs contain carefully designed distractors by domain experts, providing natural contrastive signals for learning.
Effective RLVR training should not only reward correct answers but also elicit and penalize incorrect outputs. 
Distractors can guide the model to actively distinguish among easily confusable concepts, exposing its misconceptions. 
To this end, we pose two questions: (1) How does the design of multiple-choice options affect RLVR training? (2) How can we fully exploit multiple-choice data for RLVR while preserving the format?

To address the first question, we empirically examine how candidate options affect RLVR training. By keeping the question stem constant and varying the quantity and quality of distractors, we systematically study the effect of option design. We evaluate this by measuring the model's downstream performance, examining its robustness across varying option counts and its accuracy on standard benchmarks. Through the analyses, we obtain two main findings:
(1) The impact of the number of options is overestimated. 
Although increasing options can reduce the random guess rate (25\% for 4-way vs. 10\% for 10-way questions), the effect on RLVR training is limited. 
Moreover, we observe a distributional mismatch in option counts between training and testing: inconsistency in option counts between training and test sets degrades RLVR performance. 
This suggests that in multiple-choice RLVR, the model learns patterns tied to the distribution of option counts, leading to an option-count distribution bias.
(2) Differences in distractor quality significantly influence the effectiveness of RLVR training.
Even in 2-way questions (with a 50\% chance of guessing correctly), high-quality distractors can make RLVR training effective.

For the second question, motivated by these findings, we propose \textbf{Iterative Distractor Curation (IDC)} to enhance distractor quality. 
IDC iteratively refines distractors to eliminate superficial shortcuts, compelling the model to engage in genuine reasoning rather than exploiting random guessing or trivial elimination. 
Remarkably, we find that models can self-generate these challenging options to enhance learning, effectively enabling autonomous self-improvement without any external supervision. 
We conduct experiments on four medical datasets, including MMLU-Pro Health, MedXpertQA, PubMedQA, and SuperGPQA Clinical Medicine, to validate the effectiveness of our method.

Overall, our contributions are as follows:
\begin{itemize}
    \item We show that mismatches in option count distributions between training and testing lead to performance degradation in RLVR.
    \item We analyze how option design affects RLVR and demonstrate that high-quality distractors are key to effective MCQ-based RLVR.
    \item We propose Iterative Distractor Curation (IDC) to improve MCQ distractors for RLVR training. On Qwen2-7B and Llama-3.1-8B, IDC consistently improves average accuracy: from 39.39\% to 42.62\% (+3.23\%) and from 48.63\% to 51.90\% (+3.27\%), respectively.
\end{itemize}

The remainder of this paper is organized as follows. Section~\ref{sec:related_work} reviews related work. Section~\ref{sec:preliminaries} introduces preliminaries. Section~\ref{sec:number_distractor} analyzes option quantity and validates count alignment. Section~\ref{sec:distractor_strength} investigates distractor quality and proposes IDC. Finally, Section~\ref{sec:conclusion} concludes the paper.

\section{Related Work}
\label{sec:related_work}
\paragraph{Reinforcement Learning with Verifiable Rewards (RLVR).}
RLVR~\citep{lambert2025tulu3pushingfrontiers} eliminates the reliance on human annotation in traditional RLHF~\citep{ouyang2022traininglanguagemodelsfollow} by leveraging verifiable reward signals. 
DeepSeek-R1~\citep{deepseekai2025deepseekr1incentivizingreasoningcapability, shao2024deepseekmathpushinglimitsmathematical} shows that large-scale RLVR can substantially improve LLM reasoning. 
Under this paradigm, the model samples both correct and incorrect outputs and optimizes its policy based on precise reward feedback. 
Subsequent work extends RLVR along multiple directions~\citep{zheng2025groupsequencepolicyoptimization,cui2025processreinforcementimplicitrewards,xie2025logicrlunleashingllmreasoning,zeng2025simplerlzooinvestigatingtamingzero,hu2025openreasonerzeroopensourceapproach,deepscaler2025,deepcoder2025,deepswe2025,liu2025understandingr1zeroliketrainingcritical}, but these efforts have largely focused on mathematics and coding domains, limiting their applicability to more general question-answering tasks.

To extend RLVR, recent attempts have started to utilize multiple-choice questions (MCQs)~\citep{liu2025distillationpushinglimitsmedical,zhang2025medrlvremergingmedicalreasoning,akter2025nemotroncrossthinkscalingselflearningmath,guo2025seed15vltechnicalreport,kimiteam2025kimik15scalingreinforcement}. 
The community remains divided on MCQ effectiveness: while MCQs appear to satisfy the verifiability requirement, models can guess or eliminate incorrect options to arrive at the answer, leading to unreliable reward signals. 
Seed-Thinking-v1.5~\citep{guo2025seed15vltechnicalreport} and Kimi k1.5~\cite{kimiteam2025kimik15scalingreinforcement} convert MCQs to open-ended questions or discard them, believing the MCQ format harms training.
Others use MCQs directly but report limited success. 
While~\citet{liu2025distillationpushinglimitsmedical} directly apply RLVR on original MCQs, Nemotron-CrossThink~\citep{akter2025nemotroncrossthinkscalingselflearningmath} and Med-RLVR~\citep{zhang2025medrlvremergingmedicalreasoning} observe modest gains, attributing it to restricted answer spaces. 
These approaches either abandon the verifiable MCQ format or lack systematic analysis of what affects MCQ-based RLVR effectiveness. 
In this paper, we retain the MCQ format and systematically investigate how distractor quantity and quality influence RLVR training.

\paragraph{Hard Negative Samples and Preference Learning.}
Hard negatives are foundational in contrastive learning~\cite{gao2022simcsesimplecontrastivelearning}. 
For LLM alignment, negative signals are crucial for preference learning methods like RLHF~\citep{ouyang2022traininglanguagemodelsfollow} and DPO~\citep{rafailov2024directpreferenceoptimizationlanguage}. 
Recent advances in RLVR, such as GRPO~\cite{shao2024deepseekmathpushinglimitsmathematical} and DAPO~\cite{yu2025dapoopensourcellmreinforcement}, further leverage negative samples for policy optimization.
Across these methods, negative samples typically come from two sources: (1) incorrect outputs naturally generated during training, or (2) artificially corrupted versions of correct outputs~\citep{zhang2025iopoempoweringllmscomplex}. 
Yet the role of adversarial hard negatives—specifically, MCQ distractors—remains unexplored. 
Unlike negatives generated during training, distractors are explicit, linguistically crafted hard negatives that guide models toward specific misconceptions or erroneous reasoning paths.
We demonstrate that well-designed distractors act as effective hard negatives that can substantially enhance RLVR training.

\paragraph{LLM-based Data Generation.}
LLM-based data synthesis has become mainstream~\cite{wang2024surveydatasynthesisaugmentation,wang-etal-2025-diversity,ding-etal-2024-data,luo-etal-2025-tree}, mostly focusing on generating questions (or instructions)~\citep{wang-etal-2023-self-instruct,xu2025wizardlmempoweringlargepretrained,bohnet2024longspanquestionansweringautomaticquestion}. 
In contrast, distractor generation—creating options for fixed stems—has received less attention and primarily serves evaluation benchmarks~\citep{zhang2025automatedgenerationchallengingmultiplechoice} or educational assessment~\citep{offerijns2020betterdistractionstransformerbaseddistractor}. 
Prior works evaluate distractors via extrinsic metrics such as n-gram scores (BLEU~\citep{papineni-etal-2002-bleu}, ROUGE~\citep{lin-2004-rouge}) and ranking metrics~\citep{liang-etal-2018-distractor,chiang-etal-2022-cdgp,yu-etal-2024-enhancing,2024.EDM-long-papers.1,qu-etal-2024-unsupervised}. 
The AI in Education community has also explored distractor generation for educational assessment, including LLM-based approaches~\citep{feng2024exploring,scarlatos2024improving,fernandez2024divert} and studies of alignment between LLM and student error patterns~\citep{sonkar2025imitation,liu2025natural}. 
However, the impact of distractors on LLM \emph{training}, especially RLVR, remains unexplored. 
We bridge this gap by focusing on the intrinsic training value of option design, providing both empirical evidence on how distractors affect RLVR and insights for synthesizing training-effective MCQs.

\section{Preliminaries}
\label{sec:preliminaries}

We begin by formalizing the RLVR framework for multiple-choice questions.
Consider a multiple-choice question with stem $q$ and $n$ options
$\mathcal{O} = \{o_1, \ldots, o_n\}$, where $o^*$ denotes the correct option.
The model $\pi_\theta$ generates a complete output $y = (c, a)$, where $c$ is the reasoning chain and $a \in \mathcal{O}$ is the selected answer.
The reward is defined as $r = \mathbb{I}[a = o^*]$.
The goal of RLVR is to optimize the policy $\pi_\theta$ to maximize the expected reward over the data distribution $\mathcal{D}$:
\[
\mathcal{J}(\theta) = \mathbb{E}_{(q, \mathcal{O}) \sim \mathcal{D}, y \sim \pi_\theta(\cdot \mid q, \mathcal{O})} \left[ r(y) \right].
\]
For brevity, we omit the KL divergence term. However, optimizing this objective is non-trivial. The effectiveness of RLVR critically depends on the accuracy of the reward signal~\citep{lambert2025tulu3pushingfrontiers,huang2025accuracyrobustnessstudyrule}, and prior work has shown that noisy rewards can severely degrade RL performance~\citep{shao2025spuriousrewardsrethinkingtraining,wu2025reasoningmemorizationunreliableresults}.

To analyze the reward mechanism in multiple-choice settings, we decompose the probability of receiving a reward based on reasoning validity.
We partition the space of reasoning chains into correct ($c \in \mathcal{C}_+$) and incorrect ($c \in \mathcal{C}_-$) sets.
The probabilities of receiving a reward ($r=1$) or no reward ($r=0$) can be expressed as:
\begin{align*}
\Pr[r=1] \;=\;& \Pr[a = o^* \mid c \in \mathcal{C}_+] \Pr[c \in \mathcal{C}_+] \\
&+ \Pr[a = o^* \mid c \in \mathcal{C}_-] \Pr[c \in \mathcal{C}_-], \\
\Pr[r=0] \;=\;& \Pr[a \neq o^* \mid c \in \mathcal{C}_+] \Pr[c \in \mathcal{C}_+] \\
&+ \Pr[a \neq o^* \mid c \in \mathcal{C}_-] \Pr[c \in \mathcal{C}_-].
\end{align*}
The term $\Pr[a \neq o^* \mid c \in \mathcal{C}_+]$ represents the probability that valid reasoning leads to an incorrect selection, typically due to instruction-following failures or formatting errors; assigning zero reward in these instances is consistent with our optimization goals.
However, the critical challenge arises from \textbf{spurious rewards}.
This occurs when the model selects the correct answer despite flawed reasoning:
\[
\Pr_{\text{spurious}} = \Pr[a = o^* \mid c \in \mathcal{C}_-] \Pr[c \in \mathcal{C}_-].
\]
Such rewards provide misleading positive reinforcement, encouraging the model to maintain incorrect reasoning paths.

Unlike open-ended tasks where the output space is effectively unbounded, multiple-choice questions suffer from a finite output space where incorrect reasoning is more likely to accidentally yield the correct answer. 
Although converting MCQs to open-ended formats mitigates spurious rewards, it discards distractors that embody human priors and provide natural hard negative signals. 
Additionally, rewriting risks semantic alteration or hallucination. 
Thus, we retain the original MCQ structure by fixing the stem $q$ and the correct answer $o^*$, while focusing on the distractor set $\mathcal{O}_-$ as the key control variable.
We model the probability of selecting the correct answer under incorrect reasoning as a mixture of systematic preference and random guessing:
\[
\Pr[a = o^* \mid c \in \mathcal{C}_-] = (1 - \lambda) s(\mathcal{O}_-) + \frac{\lambda}{n}.
\]
where $n$ is the total number of options, $\lambda \in [0,1]$ represents the weight of random guessing, and $s(\mathcal{O}_-) \in [0,1]$ captures the systematic preference of the model for $o^*$ over distractors despite flawed reasoning. 
This formulation reveals two key factors that control spurious rewards:
\begin{enumerate}
    \item \textbf{Number of Distractors.} The number of options determines the baseline probability of random guessing. Increasing the distractor count reduces the likelihood of accidental correctness under flawed reasoning (the $\lambda$-weighted term). This effectively mitigates spurious rewards arising from stochasticity.

    \item \textbf{Distractor Strength.} We define strength as the semantic competitiveness of distractors. \textbf{Weak distractors} are easily eliminated, allowing the model to assign high probability to $o^*$ even with flawed reasoning (increasing $s(\mathcal{O}_-)$). 
    These samples provide minimal negative feedback, and the RL algorithm fails to obtain informative penalty signals from trajectories with flawed reasoning but correct outcomes. 
    Conversely, \textbf{strong distractors} are deceptive and align with potential misconceptions. They compete for probability mass, reducing $s(\mathcal{O}_-)$ and serving as sinks for flawed reasoning. This ensures that incorrect logic leads to incorrect predictions ($r=0$).
\end{enumerate}

\section{Impact of the Number of Distractors} 
\label{sec:number_distractor}
\subsection{Analysis Study} 
\noindent\textbf{Experiment Setting.} 
We use MMLU-Pro~\citep{wang2024mmluprorobustchallengingmultitask} as the base dataset.
To construct variants with different option counts, we keep the correct answer and randomly sample distractors, yielding five variants: 2-choice (\texttt{mcq2}), 4-choice (\texttt{mcq4}), 6-choice (\texttt{mcq6}), 8-choice (\texttt{mcq8}), and 10-choice (\texttt{mcq10}). 
We train Llama-3.1-8B-Instruct and Qwen2-7B-Instruct using GRPO~\citep{shao2024deepseekmathpushinglimitsmathematical} (see Appendix~\ref{sec:hyperparameters} for hyperparameters and Appendix~\ref{sec:data-preprocessing} for data preprocessing).\footnote{We use Qwen2 instead of Qwen2.5 due to concerns about data contamination~\citep{wu2025reasoningmemorizationunreliableresults}.}

\begin{figure}[t]
    \centering 
    \includegraphics[width=\columnwidth]{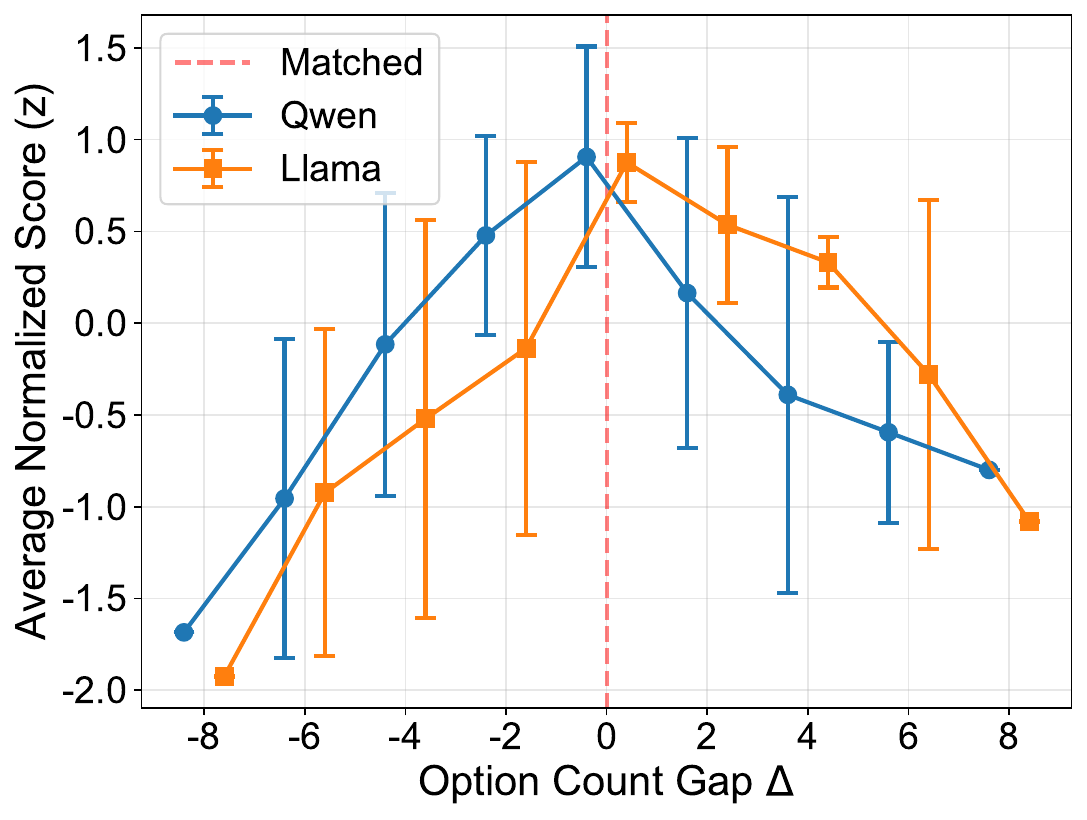}
    \caption{Relationship between average normalized $z$ score and option-count gap $\Delta$. Note that data points are slightly offset horizontally for clarity.}
    \label{fig:option_diff_comparison}
\end{figure}

\noindent\textbf{Metrics.} 
For the complete results of our cross-evaluation ($5 \text{ training} \times 5 \text{ testing}$ settings), please refer to Table~\ref{tab:full_results} in Appendix~\ref{sec:full_results_appendix}. We define the \textbf{option-count gap} $\Delta$ as:
\begin{equation}
    \Delta = m - n,
\end{equation}
where $m$ and $n$ denote the number of options in training and testing sets, respectively.
To account for the varying baseline difficulties of test sets with different option counts, we standardize performance to enable fair comparison across settings.
Specifically, we compute the normalized score:
\begin{equation}
    z_{m,n} = \frac{A_{m,n} - \mu_n}{\sigma_n},
\end{equation}
where $A_{m,n}$ denotes the accuracy of a model trained with $m$ options and evaluated on $n$ options. 
Here, $\mu_n$ and $\sigma_n$ are the mean and standard deviation of accuracy over all training option counts $m \in \{2, 4, 6, 8, 10\}$ for the fixed $n$-way test set.
Figure~\ref{fig:option_diff_comparison} illustrates the relationship between average normalized score $z$ and option-count gap $\Delta$.

\noindent\textbf{Finding: Option-count mismatch affects RLVR performance.} 
As shown in the full results (Table~\ref{tab:full_results} in Appendix~\ref{sec:full_results_appendix}), simply increasing training option count does not guarantee better performance.
Instead, we find that the alignment between training and test option counts is critical.
As shown in Figure~\ref{fig:option_diff_comparison}, three observations emerge: (1) performance peaks at $\Delta = 0$, confirming the benefit of option-count alignment; (2) performance declines monotonically as $|\Delta|$ increases, showing that larger mismatches cause greater degradation; (3) The curve exhibits asymmetry where the region with fewer training options declines more steeply than that with more options. Consequently, when $|\Delta|$ is the same, having more training options ($\Delta > 0$) is generally preferable to having fewer ($\Delta < 0$), though the advantage is model-dependent.
These findings indicate that matching training and test option counts is crucial. Larger mismatches lead to more severe performance degradation. 
A likely explanation is that the policy becomes overfitted to the specific number of choices during RLVR training. Consequently, when the option count changes at test time, the policy fails to generalize, leading to suboptimal decisions.
This phenomenon persists in larger models (Appendix~\ref{sec:mismatch_14b}), confirming it stems from a fundamental property of LLMs rather than insufficient model capacity.

\subsection{Aligning Train-Test Formats Benefits RLVR Training}

We propose that aligning the number of options between training and testing is critical for effective RLVR adaptation. We address potential mismatches as follows: if the training data has more options than the target test set, we randomly subsample the options to match the target count; if it has fewer, we augment the set by generating synthetic distractors. Specifically, the model is prompted to produce $N_{\text{test}} - N_{\text{train}}$ additional options given the question stem and existing choices. This strategy mitigates the inductive bias from mismatched formats and facilitates policy transfer.

We validate this approach on MMLU-Pro using Qwen2-7B and Llama-3.1-8B. We simulate mismatches by down-sampling original 10-choice questions to 2 or 4 choices, and then reconstructing them to 10 choices via option expansion (Table~\ref{tab:option_expansion_results}, detailed experimental setup and prompt are provided in Appendix~\ref{sec:experimental_setup_for_option_expansion}). We observe that reducing the option count gap consistently improves performance. Notably, self-generated distractors yield gains comparable to those from stronger models (e.g., GPT-OSS-120B with high reasoning effort). This result suggests that the improvement stems primarily from format alignment rather than knowledge distillation from stronger models.
However, the effectiveness varies by model. While Qwen2-7B recovers its original performance, Llama-3.1-8B lags behind the human-written baseline. This indicates that adding more options is beneficial but insufficient, suggesting that distractor quality plays a pivotal role. In the next section, we investigate the impact of distractor strength.

\begin{table}[t]
\centering
\small
\resizebox{\columnwidth}{!}{%
\begin{tabular}{l|cc}
\toprule
\textbf{Training Configuration} & \textbf{Llama-3.1-8B} & \textbf{Qwen2-7B} \\
\midrule
Train on 2-way & 52.44 & 46.66 \\
\quad 2$\to$10 (Self) & 56.12 & 50.71 \\
\quad 2$\to$10 (Qwen2.5-32B) & 55.91 & \textbf{51.01} \\
\quad 2$\to$10 (GPT-OSS-120B) & 56.26 & 50.88 \\
\midrule
Train on 4-way & 56.19 & 48.57 \\
\quad 4$\to$10 (Self) & \underline{56.72} & 50.60 \\
\quad 4$\to$10 (Qwen2.5-32B) & 56.63 & 50.59 \\
\quad 4$\to$10 (GPT-OSS-120B) & 56.44 & \underline{51.00} \\
\midrule
Train on 10-way & \textbf{58.61} & 50.07 \\
\bottomrule
\end{tabular}%
}
\caption{Effect of option expansion on RLVR performance. All results are evaluated on the \texttt{mcq10} test set and reported as accuracy (\%). \textbf{Bold} and \underline{Underlined} indicate the best and second-best results.}
\label{tab:option_expansion_results}
\end{table}

\section{The Effect of Distractor Strength}
\label{sec:distractor_strength}

\subsection{Analysis Study}
\label{sec:distractor_strength_analysis}

Based on the definition in Section~\ref{sec:preliminaries}, we hypothesize that \textit{distractor strength} determines the quality of the reward signal. Strong distractors effectively compete with the ground truth \(o^*\), providing high-contrast feedback, whereas weak distractors are trivial to reject even with flawed reasoning.
To analyze the impact of distractor strength, we conduct ablation studies by reducing MMLU-Pro questions to a 2-choice format. For a given question, we select the single distractor based on its empirical strength \(\hat{s}_j\), estimated by sampling trajectories from a reference model. Specifically, $\hat{s}_j$ is defined as the conditional frequency with which a distractor is chosen among all incorrect responses:
\begin{equation*}
\hat{s}_j =
\begin{cases}
\dfrac{\sum_{k} \mathbb{I}[a_k = o_j]}{\sum_{k} \mathbb{I}[a_k \neq o^*]}, & \text{if } \sum_{k} \mathbb{I}[a_k \neq o^*] > 0 \\
0, & \text{otherwise}
\end{cases}
\end{equation*}
where $\{a_k\}_{k=1}^K$ are answers sampled from a reference policy. This metric quantifies distractor attractiveness, where a higher $\hat{s}_j$ indicates that $o_j$ acts as a stronger trap for the model.

We analyze distractor strength by comparing three settings: (i) \textit{random}, selecting a distractor uniformly at random; (ii) \textit{strong} (\textbf{w/}), selecting the distractor with the highest \(\hat{s}_j\); and (iii) \textit{weak} (\textbf{w/o}), selecting the distractor with the lowest \(\hat{s}_j\) (detailed setup in Appendix~\ref{sec:distractor_strength_setup}). 
We experiment with Llama-3.1-8B and Qwen2-7B as policy models, using each model itself or external models (e.g., DeepSeek-V3.1) as the reference for estimating \(\hat{s}_j\).
Table~\ref{tab:distractor_strength_results} shows the experimental results.

\begin{table}[t]
\centering
\footnotesize
\setlength{\tabcolsep}{3pt}
\resizebox{\columnwidth}{!}{%
\begin{tabular}{lc|lc}
\toprule
\multicolumn{2}{c|}{\textbf{Llama-3.1-8B}} & \multicolumn{2}{c}{\textbf{Qwen2-7B}} \\
\textbf{Configuration} & \textbf{Acc} & \textbf{Configuration} & \textbf{Acc} \\
\midrule
mcq2-random & \underline{52.44} & mcq2-random & 46.66 \\
\quad w/o Llama-3.1-8B & 49.36 & \quad w/o Qwen2-7B & 39.87 \\
\quad w/ Llama-3.1-8B & \textbf{53.22} & \quad w/ Qwen2-7B & \textbf{46.98} \\
\quad w/ Qwen2-7B & 52.25 & \quad w/ Llama-3.1-8B & \underline{46.94} \\
\quad w/ DS-V3.1 & 51.13 & \quad w/ DS-V3.1 & 46.41 \\
\bottomrule
\end{tabular}%
}
\caption{Impact of distractor strength on RLVR performance (\%). We compare random distractors (mcq2-random), selecting the weakest distractor (\textbf{w/o}, lowest $\hat{s}_j$), or selecting the strongest distractor (\textbf{w/}, highest $\hat{s}_j$) as estimated by a reference model. DS-V3.1 denotes DeepSeek-V3.1. \textbf{Bold} and \underline{Underlined} indicate the best and second-best results.}
\label{tab:distractor_strength_results}
\end{table}

\noindent\textbf{Finding 1: Distractor strength is critical for RLVR training.}
We observe substantial performance differences across distractor configurations. 
For both Llama-3.1-8B and Qwen2-7B, removing strong distractors leads to performance that is clearly inferior to the random baseline. 
Conversely, reintroducing model-based strong distractors restores or improves accuracy. 
These results demonstrate that sufficiently \emph{strong} distractors are essential for learning. We attribute this to the informative preference signals they provide.

\begin{figure}[t]
    \centering
    \includegraphics[width=\columnwidth]{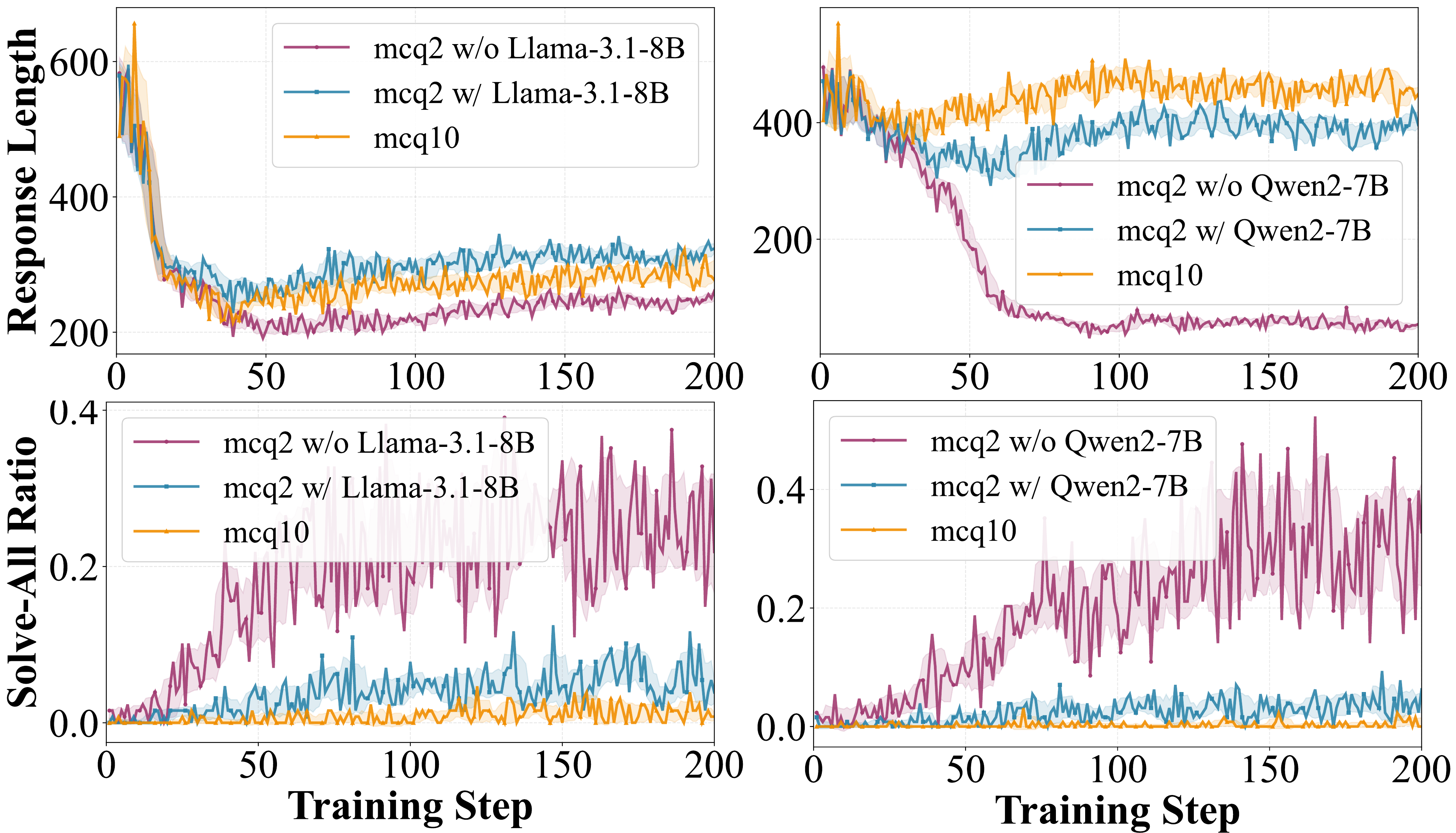}
    \caption{Training dynamics analysis: \emph{response length} (top row) and \emph{solve-all ratio} (bottom row) for \textbf{Llama} (left column) and \textbf{Qwen} (right column) across distractor settings. Removing strong distractors (w/o distractor) leads to rapid saturation and shorter responses.}
    \label{fig:solve_all_length_analysis}
\end{figure}

To further analyze this phenomenon, we visualize training dynamics in Figure~\ref{fig:solve_all_length_analysis}.
We observe that for `mcq2 w/o' (where strong distractors are removed), the solve-all ratio (the proportion of questions where the model answers correctly in all sampled attempts) increases rapidly, indicating the model quickly learns to answer many questions correctly with high probability.
In contrast, both `mcq2 w/' and `mcq10' remain challenging, maintaining a consistently lower solve-all ratio.
Additionally, `mcq2 w/o' results in shorter outputs, suggesting the model tends to skip detailed reasoning and provide answers more quickly, whereas the other two settings encourage longer responses.
These patterns hold for both Qwen and Llama.

\noindent\textbf{Finding 2: Self-targeted distractors are most effective.}
We find that distractors selected based on the target model itself yield the best results.
For both Llama and Qwen, using self-estimated distractors consistently outperforms those derived from other models (e.g., DeepSeek-V3.1) or the random baseline.
This means any model can serve as its own reference for distractor curation without relying on external models. 
Self-targeted distractors better align with the model's intrinsic error patterns and decision boundaries, providing more precise discriminative signals for optimization.

\begin{figure*}[t]
    \centering
    \includegraphics[width=\textwidth]{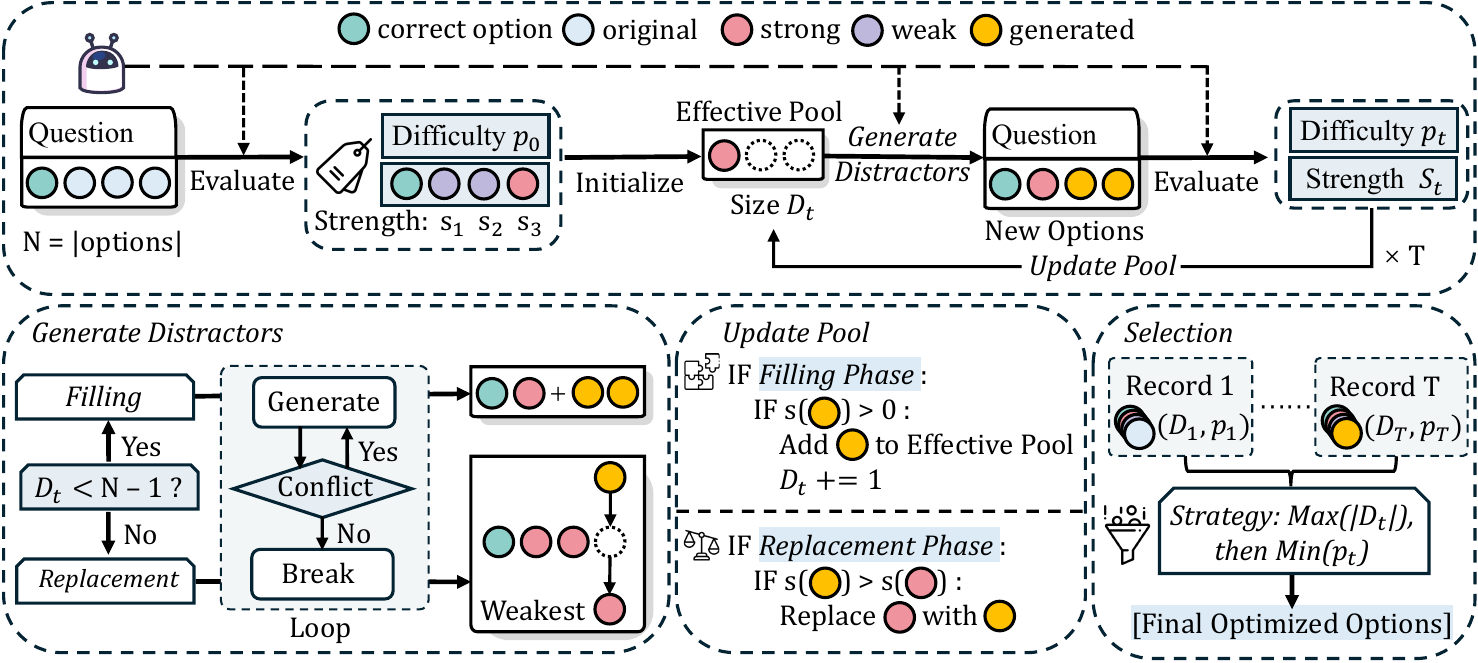}
    \caption{The framework of Iterative Distractor Curation (IDC).}
    \label{fig:idc_framework}
\end{figure*}

\begin{table*}[t]
\centering
\small
\resizebox{\textwidth}{!}{
\begin{tabular}{l|cccccc|cccccc}
\toprule
\multirow{2}{*}{\textbf{Dataset}} & \multicolumn{6}{c|}{\textbf{Qwen2-7B}} & \multicolumn{6}{c}{\textbf{Llama-3.1-8B}} \\
\cmidrule(lr){2-7} \cmidrule(lr){8-13}
 & \textbf{Base} & \textbf{Orig.} & \textbf{Direct} & \textbf{Filter} & \textbf{Rewrite} & \textbf{IDC} & \textbf{Base} & \textbf{Orig.} & \textbf{Direct} & \textbf{Filter} & \textbf{Rewrite} & \textbf{IDC} \\
\midrule
MedQA & 47.21 & 49.25 & 48.70 & 48.00 & 48.39 & \textbf{49.80} & 59.07 & 62.14 & 65.20 & 65.04 & 64.41 & \textbf{66.14} \\
SuperGPQA (Clinical) & 25.62 & 25.53 & 26.11 & 26.60 & 26.27 & \textbf{27.91} & 33.25 & 33.58 & 35.39 & 35.39 & 35.39 & \textbf{35.71} \\
MMLU-Pro (Health) & 44.38 & 46.21 & 47.07 & 46.09 & 45.72 & \textbf{49.51} & 58.56 & 57.46 & 57.95 & 59.17 & 58.19 & \textbf{60.76} \\
MedXpertQA & 10.24 & 10.98 & 10.29 & 10.86 & 10.24 & \textbf{11.88} & 14.29 & 16.29 & 15.47 & 16.45 & 15.67 & \textbf{19.31} \\
PubMedQA & 69.50 & 72.50 & 72.00 & 72.90 & 71.20 & \textbf{74.00} & 78.00 & 77.80 & 78.30 & 78.50 & \textbf{78.80} & 77.60 \\
\midrule
Average & 39.39 & 40.89 & 40.83 & 40.89 & 40.36 & \textbf{42.62} & 48.63 & 49.45 & 50.46 & 50.91 & 50.49 & \textbf{51.90} \\
\bottomrule
\end{tabular}
}
\caption{Downstream performance on medical datasets. \textbf{Base}: Base instruct model. \textbf{Orig.}: Standard RLVR on original MCQs. \textbf{Direct}: Direct conversion to short-answer. \textbf{Filter}: Filtering non-convertible questions. \textbf{Rewrite}: Model-based conversion. \textbf{IDC}: Our Iterative Distractor Curation method. All results reported as accuracy (\%).}
\label{tab:downstream_medical_results}
\end{table*}

\subsection{Iterative Distractor Curation}
\label{sec:idc}

Building on the empirical findings, 
we propose the \textbf{Iterative Distractor Curation} framework to systematically enhance distractor strength for RLVR.
We use the empirical strength $\hat{s}_j$ as the core metric to optimize the distractor set $\mathcal{O}_-(q)$ while preserving the stem and correct answer.
The process iterates through two phases:
(1) \textbf{Filling Phase}: When effective distractors (with $\hat{s}_j > 0$) are below the target count, we generate new candidates and add them if they demonstrate non-zero empirical strength.
(2) \textbf{Replacement Phase}: Once the pool reaches capacity, we identify the weakest distractor and replace it with a newly generated candidate if and only if the new candidate exhibits higher empirical strength.
After $T$ iterations, we select the option set that maximizes the number of effective distractors while minimizing the model's pass rate.
This approach functions as \emph{rejection sampling in the option space}, progressively replacing weak distractors with adversarial ones to provide clearer high-contrast reward signals.
The process is summarized in Algorithm~\ref{alg:progressive-accumulation} (see Appendix~\ref{sec:algorithm_details}), while detailed prompts for distractor generation, evaluation, and semantic equivalence checks (labeled ``Conflict'' in Figure~\ref{fig:idc_framework}) are provided in Appendix~\ref{sec:idc_prompts}.

\begin{table*}[t]
\centering
\small
\resizebox{\textwidth}{!}{
\begin{tabular}{l|cccccc|cccccc}
\toprule
\multirow{2}{*}{\textbf{Dataset}} & \multicolumn{6}{c|}{\textbf{Qwen2-7B}} & \multicolumn{6}{c}{\textbf{Llama-3.1-8B}} \\
\cmidrule(lr){2-7} \cmidrule(lr){8-13}
 & \textbf{Base} & \textbf{Self} & \textbf{Qwen-32B} & \textbf{OSS} & \textbf{DS-V3.1} & \textbf{Llama-70B} & \textbf{Base} & \textbf{Self} & \textbf{Qwen-32B} & \textbf{OSS} & \textbf{DS-V3.1} & \textbf{Llama-70B} \\
\midrule
MedQA & 47.21 & 47.29 & 49.80 & 48.70 & 49.41 & \textbf{49.88} & 59.07 & 66.61 & 66.14 & 65.75 & \textbf{67.87} & 67.09 \\
SuperGPQA & 25.62 & 27.75 & 27.91 & \textbf{28.49} & 27.75 & 27.34 & 33.25 & 36.45 & 35.71 & \textbf{38.51} & 34.81 & 36.45 \\
MMLU-Pro & 44.38 & \textbf{49.88} & 49.51 & 46.21 & 47.07 & 47.31 & 58.56 & \textbf{61.00} & 60.76 & 59.17 & 60.88 & \textbf{61.00} \\
MedXpertQA & 10.24 & 12.45 & 11.88 & \textbf{13.14} & 11.80 & 12.49 & 14.29 & \textbf{20.29} & 19.31 & 19.27 & 17.96 & 18.04 \\
PubMedQA & 69.50 & \textbf{74.60} & 74.00 & 71.60 & 71.50 & 72.50 & 78.00 & \textbf{78.10} & 77.60 & 76.80 & 77.90 & 77.00 \\
\midrule
\textbf{Average} & 39.39 & 42.39 & \textbf{42.62} & 41.63 & 41.51 & 41.90 & 48.63 & \textbf{52.49} & 51.90 & 51.90 & 51.88 & 51.92 \\
\bottomrule
\end{tabular}
}
\caption{Ablation on distractor generators. We compare models trained with distractors generated by the model itself (``Self'') versus stronger external models. ``Base'' denotes the original instruct model without RLVR. Qwen-32B: Qwen2.5-32B-Instruct; OSS: GPT-OSS-120B; DS-V3.1: DeepSeek-V3.1; Llama-70B: Llama-3.1-70B-Instruct.}
\label{tab:generator_ablation}
\end{table*}

\subsection{Experimental Setup}
\label{sec:experimental_setup}

In this section, we apply the proposed framework to different models and evaluate its impact on downstream tasks. 
We use the following experimental settings (see Appendix~\ref{sec:hyperparameters} for details).

\noindent\textbf{Models.}
Following the analytical experiments, we use Qwen2-7B-Instruct and Llama-3.1-8B-Instruct for both RLVR training and distractor strength estimation.
For distractor generation, we employ Qwen2.5-32B-Instruct as the default generator due to its strong instruction-following ability and moderate cost.
We further investigate the impact of generator models and explore the potential for \textbf{self-improvement} (using the target model itself as the generator) in the ablation studies.

\noindent\textbf{Datasets.}
We focus on the medical domain.
We use the MedQA~\cite{medqa} training split as our training data, and evaluate in-domain performance on the MedQA test set.
For out-of-distribution (OOD) evaluation, we use the clinical subset of SuperGPQA~\cite{pteam2025supergpqascalingllmevaluation} and the health subset of MMLU-Pro~\cite{wang2024mmluprorobustchallengingmultitask}. 
We also evaluate on MedXpertQA~\cite{MedXpertQA}, a challenging benchmark designed for evaluating expert-level medical knowledge and reasoning capabilities.
In addition, we evaluate the trained models on PubMedQA~\cite{jin-etal-2019-pubmedqa} to measure transfer to the short-answer (QA) format.

\noindent\textbf{Baselines.}
We compare our method against the following baselines (see Appendix~\ref{sec:baseline_prompts} for details):
\begin{itemize}
    \item \emph{Direct conversion:} strip options and convert all questions to a short-answer format.
    \item \emph{Filtering non-convertible questions~\citep{akter2025nemotroncrossthinkscalingselflearningmath}:} first filter out multiple-choice questions that cannot be directly converted, then convert remaining questions to short-answer questions by stripping the options.
    \item \emph{Model-based conversion:} use a language model to rewrite each question into a fill-in-the-blank or short-answer format.
\end{itemize}

\subsection{Results}

Table~\ref{tab:downstream_medical_results} presents the results on downstream medical tasks. IDC improves performance in the medical domain on both in-domain and OOD evaluations, consistently outperforming baselines based on direct conversion or model-based rewriting. The only exception is PubMedQA with Llama-3.1-8B, which we attribute to format alignment bias: PubMedQA is a short-answer QA task (not MCQ), so conversion baselines enjoy a direct format-alignment advantage on this benchmark. This effect is amplified by Llama's stronger sensitivity to format mismatch---as Table~\ref{tab:full_results} shows, Llama exhibits strict diagonal dominance in cross-option generalization. On Qwen2-7B, which is less format-sensitive, IDC outperforms all baselines on PubMedQA, confirming that IDC's gains do transfer beyond format boundaries.
Multi-seed experiments (Appendix~\ref{sec:multi_seed}) confirm that these gains are stable across independent runs, and an additional law domain experiment (Appendix~\ref{sec:law_domain}) validates that IDC generalizes beyond the medical domain.

\subsection{Ablation Studies}
\label{sec:ablation_studies}

\begin{figure}[t]
    \centering
    \includegraphics[width=\columnwidth]{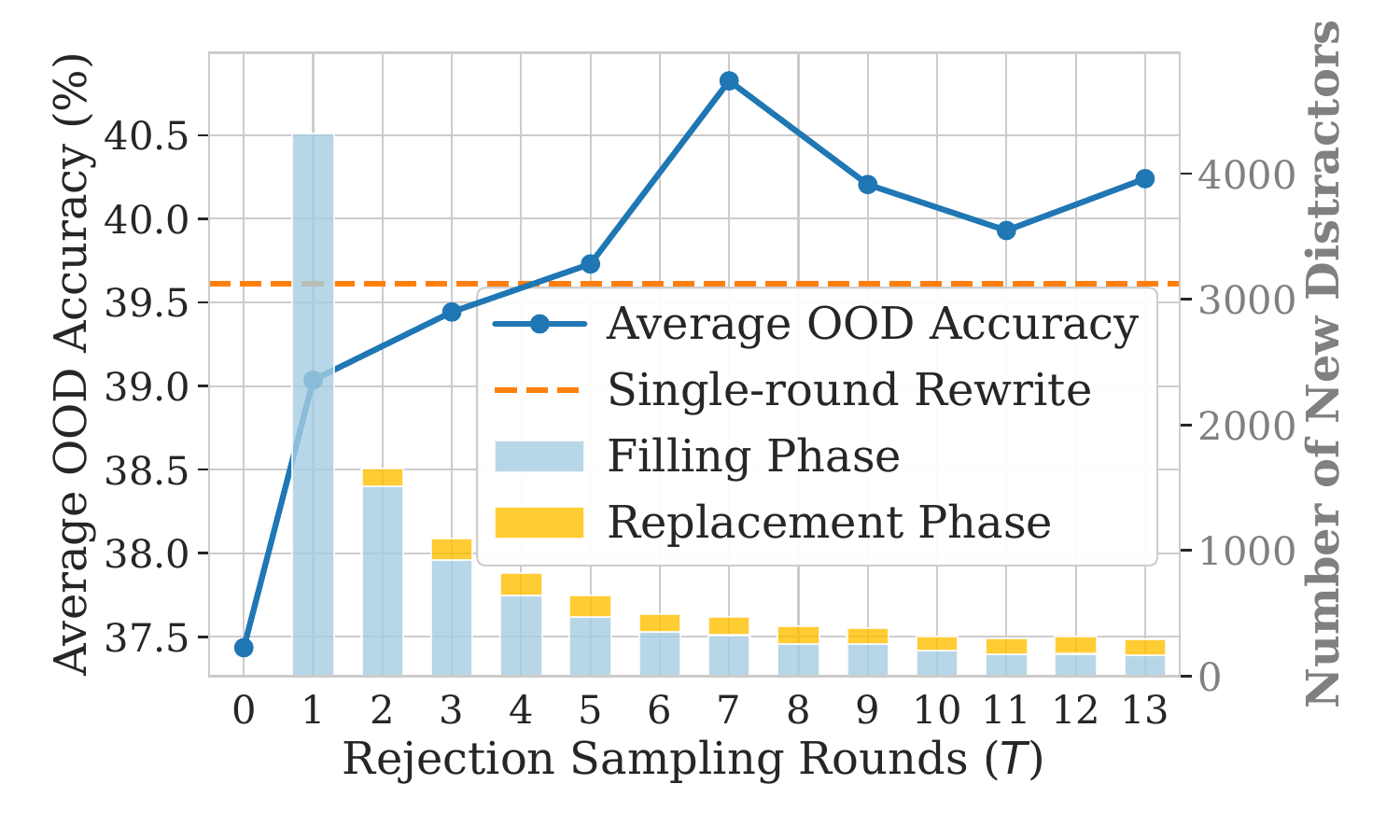}
    \caption{Performance gains and distractor dynamics across rejection sampling rounds.}
    \label{fig:rs_rounds}
\end{figure}

\noindent\textbf{Impact of rejection sampling rounds.}
We study the influence of rejection sampling iterations $T$.
Figure~\ref{fig:rs_rounds} compares our iterative approach with a single-round baseline that rewrites all distractors at once without strength estimation. Note that the training data (MedQA) uses 4-choice questions (1 correct + 3 distractors), so the single-round baseline replaces all 3 distractors simultaneously, while IDC replaces one at a time.
The single-round strategy (dashed line) is outperformed by our iterative method, demonstrating 
the necessity of iterative refinement.
We observe that newly effective distractors decrease rapidly after the initial rounds.
Based on this saturation, we fix $T=7$ for all experiments 
to balance performance and efficiency.
Experimental details are in Appendix~\ref{sec:rs_setup}.

\noindent\textbf{Computational cost analysis.}
Table~\ref{tab:cost_breakdown} summarizes the wall-clock time for each stage. The overhead of IDC is modest: strength evaluation only requires the model to output a single answer label without chain-of-thought, and the target model itself serves as both generator and evaluator (Table~\ref{tab:generator_ablation}), eliminating the need for any external model. The full IDC pipeline accounts for less than 10\% of the total pipeline time.

\begin{table}[t]
\centering
\small
\begin{tabular}{l|r}
\toprule
\textbf{Stage} & \textbf{Wall-Clock Time} \\
\midrule
Distractor Generation (per round) & $\sim$15--20 min \\
Strength Evaluation (per round) & $<$5 min \\
Full IDC Pipeline (7 rounds) & $\sim$1.5 h \\
RLVR Training & 18--20 h \\
\midrule
\textbf{IDC Overhead / Total} & \textbf{$<$10\%} \\
\bottomrule
\end{tabular}
\caption{Computational cost breakdown on 8$\times$H200 GPUs (MedQA, $\sim$10k samples).}
\label{tab:cost_breakdown}
\end{table}

\noindent\textbf{Spurious reward rate analysis.}
To empirically validate our theoretical framework (Section~\ref{sec:preliminaries}), we measure how distractor design affects spurious reward rates (Table~\ref{tab:spurious_reward}). For each sampled response, DeepSeek-V3.1 provides structured judgments on answer correctness, reasoning correctness, and whether a spurious reward occurred (correct answer despite flawed reasoning), followed by human verification. Results are pooled over Qwen2-7B and Llama-3.1-8B. Both increasing option count and applying IDC substantially reduce spurious rewards, confirming that option design directly impacts reward signal accuracy.

\begin{table}[t]
\centering
\small
\setlength{\tabcolsep}{5pt}
\begin{tabular}{cc c cc}
\toprule
\multirow{2}{*}{\textbf{Factor}} & \multirow{2}{*}{\textbf{Setting}} & \multirow{2}{*}{$\mathbf{n}$} & \multicolumn{2}{c}{\textbf{Spurious Rate}} \\
\cmidrule(lr){4-5}
 & & & Before & After \\
\midrule
Quantity & \texttt{mcq2} $\rightarrow$ \texttt{mcq10} & 400 & 0.241 & \textbf{0.095} \\
Quality & w/o $\rightarrow$ w/ IDC & 200 & 0.260 & \textbf{0.110} \\
\bottomrule
\end{tabular}
\caption{Spurious reward rates before and after varying option design on MMLU-Pro (row~1) and MedQA (row~2). $n$: annotated response pairs.}
\label{tab:spurious_reward}
\end{table}

\noindent\textbf{Ablation Study on Distractor Generators}
We investigate whether a stronger generator necessarily leads to better performance.
Table~\ref{tab:generator_ablation} compares using the target model itself against stronger external models as generators.
Results show that stronger generators do not guarantee better training outcomes.
Self-generated distractors yield competitive or superior results on MMLU-Pro and PubMedQA.
This suggests that distractor effectiveness depends on the current policy: self-generated options likely lie closer to the model's decision boundary, providing more pertinent supervision than ``harder'' but distributionally distinct options from stronger models.
Additionally, the experiment in Appendix~\ref{sec:qwen14b_ablation} demonstrates the scalability of our approach to larger model architectures.
We also verify in Appendix~\ref{sec:estimator_ablation} that self-verification (estimating strength with the model itself) is sufficient.

\noindent\textbf{Semantic collapse analysis.}
A potential failure mode of self-generated distractors is semantic collapse, where a generated distractor becomes semantically equivalent to the correct answer. We analyze this on 5,000 MedQA training questions (35,000 IDC generation rounds). Among all candidates, 652 were identified as equivalent to the gold answer (648 semantically, 4 exact matches), spanning 222 out of 5,000 questions (4.44\%). Representative cases include medical aliases: \textit{Vitamin B1} vs.\ \textit{Thiamine}, \textit{Giant cell arteritis} vs.\ \textit{Temporal arteritis}, and \textit{Anti-Ro antibodies} vs.\ \textit{Anti-SSA antibodies}. Such cases are caught by the semantic equivalence check (Section~\ref{sec:idc}) and replaced with new candidates, ensuring reward signal integrity.

% \noindent\textbf{Case study.}
% Appendix~\ref{sec:case_study} illustrates how questions evolve across rejection-sampling rounds, with options becoming progressively more challenging.

\section{Conclusions}
\label{sec:conclusion}
In this work, we systematically study factors influencing the suitability of multiple-choice questions for RLVR. Through controlled experiments on MMLU-Pro, we find that train-test consistency in option count and distractor strength are key factors. Based on these insights, we propose model-based option expansion and strength-guided distractor synthesis. This approach addresses the identified issues and enhances utilization of existing multiple-choice data. 
Crucially, models can self-generate these challenging distractors, enabling self-improvement without external supervision.

\section*{Limitations}
Our method effectively mitigates challenges in applying RLVR to multiple-choice questions. However, predicting a model's capability to synthesize effective distractors remains difficult \textit{a priori}. The relationship between general model capability and synthetic distractor quality is not yet clear. We believe that training specialized distractor generators aligned with RLVR objectives is an important direction for future work.

\section*{Use of AI Assistants}

We primarily use AI assistants to improve and enrich our writing, especially by leveraging LLMs to help us write taxonomy in \LaTeX.

% \section*{Acknowledgments}

\bibliography{references-used}

@misc{lambert2025tulu3pushingfrontiers,
  author = {Nathan Lambert and Jacob Morrison and
            Valentina Pyatkin and Shengyi Huang and Hamish Ivison and
            Faeze Brahman and Lester James V. Miranda and
            Alisa Liu and Nouha Dziri and Shane Lyu and Yuling Gu and
            Saumya Malik and Victoria Graf and Jena D. Hwang and
            Jiangjiang Yang and Ronan Le Bras and Oyvind Tafjord and
            Chris Wilhelm and Luca Soldaini and Noah A. Smith and
            Yizhong Wang and Pradeep Dasigi and
            Hannaneh Hajishirzi},
  title  = {Tulu 3: Pushing Frontiers in Open Language Model
            Post-Training},
  year   = {2025},
  url    = {https://arxiv.org/abs/2411.15124}
}

@misc{shao2024deepseekmathpushinglimitsmathematical,
  author = {Zhihong Shao and Peiyi Wang and Qihao Zhu and
            Runxin Xu and Junxiao Song and Xiao Bi and
            Haowei Zhang and Mingchuan Zhang and Y. K. Li and
            Y. Wu and Daya Guo},
  title  = {DeepSeekMath: Pushing the Limits of Mathematical
            Reasoning in Open Language Models},
  year   = {2024},
  url    = {https://arxiv.org/abs/2402.03300}
}

@misc{chen2025surveyinductivereasoninglarge,
  author = {Kedi Chen and Dezhao Ruan and Yuhao Dan and
            Yaoting Wang and Siyu Yan and Xuecheng Wu and
            Yinqi Zhang and Qin Chen and Jie Zhou and Liang He and
            Biqing Qi and Linyang Li and Qipeng Guo and
            Xiaoming Shi and Wei Zhang},
  title  = {A Survey of Inductive Reasoning for Large Language
            Models},
  year   = {2025},
  url    = {https://arxiv.org/abs/2510.10182}
}

@misc{ouyang2022traininglanguagemodelsfollow,
  author = {Long Ouyang and Jeff Wu and Xu Jiang and
            Diogo Almeida and Carroll L. Wainwright and
            Pamela Mishkin and Chong Zhang and Sandhini Agarwal and
            Katarina Slama and Alex Ray and John Schulman and
            Jacob Hilton and Fraser Kelton and Luke Miller and
            Maddie Simens and Amanda Askell and Peter Welinder and
            Paul Christiano and Jan Leike and Ryan Lowe},
  title  = {Training language models to follow instructions with
            human feedback},
  year   = {2022},
  url    = {https://arxiv.org/abs/2203.02155}
}

@misc{liu2025distillationpushinglimitsmedical,
  author = {Che Liu and Haozhe Wang and Jiazhen Pan and
            Zhongwei Wan and Yong Dai and Fangzhen Lin and
            Wenjia Bai and Daniel Rueckert and Rossella Arcucci},
  title  = {Beyond Distillation: Pushing the Limits of Medical
            LLM Reasoning with Minimalist Rule-Based RL},
  year   = {2025},
  url    = {https://arxiv.org/abs/2505.17952}
}

@misc{zhang2025medrlvremergingmedicalreasoning,
  author = {Sheng Zhang and Qianchu Liu and Guanghui Qin and
            Tristan Naumann and Hoifung Poon},
  title  = {Med-RLVR: Emerging Medical Reasoning from a 3B base
            model via reinforcement Learning},
  year   = {2025},
  url    = {https://arxiv.org/abs/2502.19655}
}

@misc{akter2025nemotroncrossthinkscalingselflearningmath,
  author = {Syeda Nahida Akter and Shrimai Prabhumoye and
            Matvei Novikov and Seungju Han and Ying Lin and
            Evelina Bakhturina and Eric Nyberg and Yejin Choi and
            Mostofa Patwary and Mohammad Shoeybi and
            Bryan Catanzaro},
  title  = {Nemotron-CrossThink: Scaling Self-Learning beyond
            Math Reasoning},
  year   = {2025},
  url    = {https://arxiv.org/abs/2504.13941}
}

@misc{guo2025seed15vltechnicalreport,
  author = {Dong Guo and Faming Wu and Feida Zhu and Fuxing Leng and
            Guang Shi and Haobin Chen and others},
  title  = {Seed1.5-VL Technical Report},
  year   = {2025},
  url    = {https://arxiv.org/abs/2505.07062}
}

@misc{amodei2016concreteproblemsaisafety,
  author = {Dario Amodei and Chris Olah and Jacob Steinhardt and
            Paul Christiano and John Schulman and Dan Mané},
  title  = {Concrete Problems in AI Safety},
  year   = {2016},
  url    = {https://arxiv.org/abs/1606.06565}
}

@misc{kimiteam2025kimik15scalingreinforcement,
  author = {Kimi Team and Angang Du and Bofei Gao and Bowei Xing and
            Changjiu Jiang and others},
  title  = {Kimi k1.5: Scaling Reinforcement Learning with LLMs},
  year   = {2025},
  url    = {https://arxiv.org/abs/2501.12599}
}

@misc{deepseekai2025deepseekr1incentivizingreasoningcapability,
  author = {DeepSeek-AI and Daya Guo and Dejian Yang and
            Haowei Zhang and Junxiao Song and Ruoyu Zhang and
            Runxin Xu and Qihao Zhu and Shirong Ma and Peiyi Wang and
            Xiao Bi and Xiaokang Zhang and Xingkai Yu and Yu Wu and
            Z. F. Wu and Zhibin Gou and Zhihong Shao and
            Zhuoshu Li and Ziyi Gao and Aixin Liu and Bing Xue and
            Bingxuan Wang and Bochao Wu and Bei Feng and
            Chengda Lu and Chenggang Zhao and Chengqi Deng and
            Chenyu Zhang and Chong Ruan and Damai Dai and
            Deli Chen and Dongjie Ji and Erhang Li and
            Fangyun Lin and Fucong Dai and Fuli Luo and
            Guangbo Hao and Guanting Chen and Guowei Li and
            H. Zhang and Han Bao and Hanwei Xu and Haocheng Wang and
            Honghui Ding and Huajian Xin and Huazuo Gao and
            Hui Qu and Hui Li and Jianzhong Guo and Jiashi Li and
            Jiawei Wang and Jingchang Chen and Jingyang Yuan and
            Junjie Qiu and Junlong Li and J. L. Cai and Jiaqi Ni and
            Jian Liang and Jin Chen and Kai Dong and Kai Hu and
            Kaige Gao and Kang Guan and Kexin Huang and Kuai Yu and
            Lean Wang and Lecong Zhang and Liang Zhao and
            Litong Wang and Liyue Zhang and Lei Xu and Leyi Xia and
            Mingchuan Zhang and Minghua Zhang and Minghui Tang and
            Meng Li and Miaojun Wang and Mingming Li and
            Ning Tian and Panpan Huang and Peng Zhang and
            Qiancheng Wang and Qinyu Chen and Qiushi Du and
            Ruiqi Ge and Ruisong Zhang and Ruizhe Pan and
            Runji Wang and R. J. Chen and R. L. Jin and Ruyi Chen and
            Shanghao Lu and Shangyan Zhou and Shanhuang Chen and
            Shengfeng Ye and Shiyu Wang and Shuiping Yu and
            Shunfeng Zhou and Shuting Pan and S. S. Li and
            Shuang Zhou and Shaoqing Wu and Shengfeng Ye and
            Tao Yun and Tian Pei and Tianyu Sun and T. Wang and
            Wangding Zeng and Wanjia Zhao and Wen Liu and
            Wenfeng Liang and Wenjun Gao and Wenqin Yu and
            Wentao Zhang and W. L. Xiao and Wei An and
            Xiaodong Liu and Xiaohan Wang and Xiaokang Chen and
            Xiaotao Nie and Xin Cheng and Xin Liu and Xin Xie and
            Xingchao Liu and Xinyu Yang and Xinyuan Li and
            Xuecheng Su and Xuheng Lin and X. Q. Li and
            Xiangyue Jin and Xiaojin Shen and Xiaosha Chen and
            Xiaowen Sun and Xiaoxiang Wang and Xinnan Song and
            Xinyi Zhou and Xianzu Wang and Xinxia Shan and
            Y. K. Li and Y. Q. Wang and Y. X. Wei and Yang Zhang and
            Yanhong Xu and Yao Li and Yao Zhao and Yaofeng Sun and
            Yaohui Wang and Yi Yu and Yichao Zhang and Yifan Shi and
            Yiliang Xiong and Ying He and Yishi Piao and
            Yisong Wang and Yixuan Tan and Yiyang Ma and
            Yiyuan Liu and Yongqiang Guo and Yuan Ou and
            Yuduan Wang and Yue Gong and Yuheng Zou and Yujia He and
            Yunfan Xiong and Yuxiang Luo and Yuxiang You and
            Yuxuan Liu and Yuyang Zhou and Y. X. Zhu and
            Yanhong Xu and Yanping Huang and Yaohui Li and
            Yi Zheng and Yuchen Zhu and Yunxian Ma and Ying Tang and
            Yukun Zha and Yuting Yan and Z. Z. Ren and Zehui Ren and
            Zhangli Sha and Zhe Fu and Zhean Xu and Zhenda Xie and
            Zhengyan Zhang and Zhewen Hao and Zhicheng Ma and
            Zhigang Yan and Zhiyu Wu and Zihui Gu and Zijia Zhu and
            Zijun Liu and Zilin Li and Ziwei Xie and Ziyang Song and
            Zizheng Pan and Zhen Huang and Zhipeng Xu and
            Zhongyu Zhang and Zhen Zhang},
  title  = {DeepSeek-R1: Incentivizing Reasoning Capability in
            LLMs via Reinforcement Learning},
  year   = {2025},
  url    = {https://arxiv.org/abs/2501.12948}
}

@misc{zheng2025groupsequencepolicyoptimization,
  author = {Chujie Zheng and Shixuan Liu and Mingze Li and
            Xiong-Hui Chen and Bowen Yu and Chang Gao and
            Kai Dang and Yuqiong Liu and Rui Men and An Yang and
            Jingren Zhou and Junyang Lin},
  title  = {Group Sequence Policy Optimization},
  year   = {2025},
  url    = {https://arxiv.org/abs/2507.18071}
}

@misc{cui2025processreinforcementimplicitrewards,
  author = {Ganqu Cui and Lifan Yuan and Zefan Wang and
            Hanbin Wang and Yuchen Zhang and Jiacheng Chen and
            Wendi Li and Bingxiang He and Yuchen Fan and
            Tianyu Yu and Qixin Xu and Weize Chen and Jiarui Yuan and
            Huayu Chen and Kaiyan Zhang and Xingtai Lv and
            Shuo Wang and Yuan Yao and Xu Han and Hao Peng and
            Yu Cheng and Zhiyuan Liu and Maosong Sun and
            Bowen Zhou and Ning Ding},
  title  = {Process Reinforcement through Implicit Rewards},
  year   = {2025},
  url    = {https://arxiv.org/abs/2502.01456}
}

@misc{xie2025logicrlunleashingllmreasoning,
  author = {Tian Xie and Zitian Gao and Qingnan Ren and
            Haoming Luo and Yuqian Hong and Bryan Dai and
            Joey Zhou and Kai Qiu and Zhirong Wu and Chong Luo},
  title  = {Logic-RL: Unleashing LLM Reasoning with Rule-Based
            Reinforcement Learning},
  year   = {2025},
  url    = {https://arxiv.org/abs/2502.14768}
}

@misc{zeng2025simplerlzooinvestigatingtamingzero,
  author = {Weihao Zeng and Yuzhen Huang and Qian Liu and Wei Liu and
            Keqing He and Zejun Ma and Junxian He},
  title  = {SimpleRL-Zoo: Investigating and Taming Zero
            Reinforcement Learning for Open Base Models in the
            Wild},
  year   = {2025},
  url    = {https://arxiv.org/abs/2503.18892}
}

@misc{hu2025openreasonerzeroopensourceapproach,
  author = {Jingcheng Hu and Yinmin Zhang and Qi Han and
            Daxin Jiang and Xiangyu Zhang and Heung-Yeung Shum},
  title  = {Open-Reasoner-Zero: An Open Source Approach to
            Scaling Up Reinforcement Learning on the Base Model},
  year   = {2025},
  url    = {https://arxiv.org/abs/2503.24290}
}

@misc{deepscaler2025,
  author       = {Michael Luo and Sijun Tan and Justin Wong and
                  Xiaoxiang Shi and William Y. Tang and Manan Roongta and
                  Colin Cai and Jeffrey Luo and Li Erran Li and
                  Raluca Ada Popa and Ion Stoica},
  howpublished = {\url{https://pretty-radio-b75.notion.site/DeepScaleR-Surpassing-O1-Preview-with-a-1-5B-Model-by-Scaling-RL-19681902c1468005bed8ca303013a4e2}},
  note         = {Notion Blog},
  title        = {DeepScaleR: Surpassing O1-Preview with a 1.5B Model
                  by Scaling RL},
  year         = {2025}
}

@misc{deepcoder2025,
  author       = {Michael Luo and Sijun Tan and Roy Huang and
                  Ameen Patel and Alpay Ariyak and Qingyang Wu and
                  Xiaoxiang Shi and Rachel Xin and Colin Cai and
                  Maurice Weber and Ce Zhang and Li Erran Li and
                  Raluca Ada Popa and Ion Stoica},
  howpublished = {\url{https://pretty-radio-b75.notion.site/DeepCoder-A-Fully-Open-Source-14B-Coder-at-O3-mini-Level-1cf81902c14680b3bee5eb349a512a51}},
  note         = {Notion Blog},
  title        = {DeepCoder: A Fully Open-Source 14B Coder at O3-mini
                  Level},
  year         = {2025}
}

@misc{deepswe2025,
  author       = {Michael Luo and Naman Jain and Jaskirat Singh and
                  Sijun Tan and Ameen Patel and Qingyang Wu and
                  Alpay Ariyak and Colin Cai and Tarun Venkat and
                  Shang Zhu and Ben Athiwaratkun and Manan Roongta and
                  Ce Zhang and Li Erran Li and Raluca Ada Popa and
                  Koushik Sen and Ion Stoica},
  howpublished = {\url{https://pretty-radio-b75.notion.site/DeepSWE-Training-a-Fully-Open-sourced-State-of-the-Art-Coding-Agent-by-Scaling-RL-22281902c1468193aabbe9a8c59bbe33}},
  note         = {Notion Blog},
  title        = {DeepSWE: Training a State-of-the-Art Coding Agent
                  from Scratch by Scaling RL},
  year         = {2025}
}

@misc{liu2025understandingr1zeroliketrainingcritical,
  author = {Zichen Liu and Changyu Chen and Wenjun Li and
            Penghui Qi and Tianyu Pang and Chao Du and
            Wee Sun Lee and Min Lin},
  title  = {Understanding R1-Zero-Like Training: A Critical
            Perspective},
  year   = {2025},
  url    = {https://arxiv.org/abs/2503.20783}
}

@misc{gao2022simcsesimplecontrastivelearning,
  author = {Tianyu Gao and Xingcheng Yao and Danqi Chen},
  title  = {SimCSE: Simple Contrastive Learning of Sentence
            Embeddings},
  year   = {2022},
  url    = {https://arxiv.org/abs/2104.08821}
}

@misc{rafailov2024directpreferenceoptimizationlanguage,
  author = {Rafael Rafailov and Archit Sharma and Eric Mitchell and
            Stefano Ermon and Christopher D. Manning and
            Chelsea Finn},
  title  = {Direct Preference Optimization: Your Language Model
            is Secretly a Reward Model},
  year   = {2024},
  url    = {https://arxiv.org/abs/2305.18290}
}

@misc{yu2025dapoopensourcellmreinforcement,
  author = {Qiying Yu and Zheng Zhang and Ruofei Zhu and
            Yufeng Yuan and Xiaochen Zuo and Yu Yue and
            Weinan Dai and Tiantian Fan and Gaohong Liu and
            Lingjun Liu and Xin Liu and Haibin Lin and Zhiqi Lin and
            Bole Ma and Guangming Sheng and Yuxuan Tong and
            Chi Zhang and Mofan Zhang and Wang Zhang and Hang Zhu and
            Jinhua Zhu and Jiaze Chen and Jiangjie Chen and
            Chengyi Wang and Hongli Yu and Yuxuan Song and
            Xiangpeng Wei and Hao Zhou and Jingjing Liu and
            Wei-Ying Ma and Ya-Qin Zhang and Lin Yan and Mu Qiao and
            Yonghui Wu and Mingxuan Wang},
  title  = {DAPO: An Open-Source LLM Reinforcement Learning
            System at Scale},
  year   = {2025},
  url    = {https://arxiv.org/abs/2503.14476}
}

@misc{zhang2025iopoempoweringllmscomplex,
  author = {Xinghua Zhang and Haiyang Yu and Cheng Fu and
            Fei Huang and Yongbin Li},
  title  = {IOPO: Empowering LLMs with Complex Instruction
            Following via Input-Output Preference Optimization},
  year   = {2025},
  url    = {https://arxiv.org/abs/2411.06208}
}

@misc{wang2024surveydatasynthesisaugmentation,
  author = {Ke Wang and Jiahui Zhu and Minjie Ren and Zeming Liu and
            Shiwei Li and Zongye Zhang and Chenkai Zhang and
            Xiaoyu Wu and Qiqi Zhan and Qingjie Liu and
            Yunhong Wang},
  title  = {A Survey on Data Synthesis and Augmentation for Large
            Language Models},
  year   = {2024},
  url    = {https://arxiv.org/abs/2410.12896}
}

@inproceedings{wang-etal-2025-diversity,
  address   = {Vienna, Austria},
  author    = {Wang, Zaitian and Zhang, Jinghan and Zhang, Xinhao and
               Liu, Kunpeng and Wang, Pengfei and Zhou, Yuanchun},
  booktitle = {Proceedings of the 63rd Annual Meeting of the
               Association for Computational Linguistics (Volume 1:
               Long Papers)},
  editor    = {Che, Wanxiang and Nabende, Joyce and
               Shutova, Ekaterina and Pilehvar, Mohammad Taher},
  month     = jul,
  pages     = {22265--22283},
  publisher = {Association for Computational Linguistics},
  title     = {Diversity-oriented Data Augmentation with Large
               Language Models},
  year      = {2025},
  doi       = {10.18653/v1/2025.acl-long.1084},
  isbn      = {979-8-89176-251-0},
  url       = {https://aclanthology.org/2025.acl-long.1084/}
}

@inproceedings{ding-etal-2024-data,
  address   = {Bangkok, Thailand},
  author    = {Ding, Bosheng and Qin, Chengwei and Zhao, Ruochen and
               Luo, Tianze and Li, Xinze and Chen, Guizhen and
               Xia, Wenhan and Hu, Junjie and Luu, Anh Tuan and
               Joty, Shafiq},
  booktitle = {Findings of the Association for Computational
               Linguistics: ACL 2024},
  editor    = {Ku, Lun-Wei and Martins, Andre and Srikumar, Vivek},
  month     = aug,
  pages     = {1679--1705},
  publisher = {Association for Computational Linguistics},
  title     = {Data Augmentation using {LLM}s: Data Perspectives,
               Learning Paradigms and Challenges},
  year      = {2024},
  doi       = {10.18653/v1/2024.findings-acl.97},
  url       = {https://aclanthology.org/2024.findings-acl.97/}
}

@inproceedings{luo-etal-2025-tree,
  address   = {Vienna, Austria},
  author    = {Luo, Ziyang and Li, Kaixin and Lin, Hongzhan and
               Tian, Yuchen and Kankanhalli, Mohan and Ma, Jing},
  booktitle = {Proceedings of the 63rd Annual Meeting of the
               Association for Computational Linguistics (Volume 1:
               Long Papers)},
  editor    = {Che, Wanxiang and Nabende, Joyce and
               Shutova, Ekaterina and Pilehvar, Mohammad Taher},
  month     = jul,
  pages     = {297--316},
  publisher = {Association for Computational Linguistics},
  title     = {Tree-of-Evolution: Tree-Structured Instruction
               Evolution for Code Generation in Large Language
               Models},
  year      = {2025},
  doi       = {10.18653/v1/2025.acl-long.14},
  isbn      = {979-8-89176-251-0},
  url       = {https://aclanthology.org/2025.acl-long.14/}
}

@inproceedings{wang-etal-2023-self-instruct,
  address   = {Toronto, Canada},
  author    = {Wang, Yizhong and Kordi, Yeganeh and Mishra, Swaroop and
               Liu, Alisa and Smith, Noah A. and Khashabi, Daniel and
               Hajishirzi, Hannaneh},
  booktitle = {Proceedings of the 61st Annual Meeting of the
               Association for Computational Linguistics (Volume 1:
               Long Papers)},
  editor    = {Rogers, Anna and Boyd-Graber, Jordan and
               Okazaki, Naoaki},
  month     = jul,
  pages     = {13484--13508},
  publisher = {Association for Computational Linguistics},
  title     = {Self-Instruct: Aligning Language Models with
               Self-Generated Instructions},
  year      = {2023},
  doi       = {10.18653/v1/2023.acl-long.754},
  url       = {https://aclanthology.org/2023.acl-long.754/}
}

@misc{xu2025wizardlmempoweringlargepretrained,
  author = {Can Xu and Qingfeng Sun and Kai Zheng and Xiubo Geng and
            Pu Zhao and Jiazhan Feng and Chongyang Tao and
            Qingwei Lin and Daxin Jiang},
  title  = {WizardLM: Empowering large pre-trained language
            models to follow complex instructions},
  year   = {2025},
  url    = {https://arxiv.org/abs/2304.12244}
}

@misc{bohnet2024longspanquestionansweringautomaticquestion,
  author = {Bernd Bohnet and Kevin Swersky and Rosanne Liu and
            Pranjal Awasthi and Azade Nova and Javier Snaider and
            Hanie Sedghi and Aaron T Parisi and Michael Collins and
            Angeliki Lazaridou and Orhan Firat and Noah Fiedel},
  title  = {Long-Span Question-Answering: Automatic Question
            Generation and QA-System Ranking via Side-by-Side
            Evaluation},
  year   = {2024},
  url    = {https://arxiv.org/abs/2406.00179}
}

@misc{zhang2025automatedgenerationchallengingmultiplechoice,
  author = {Yuhui Zhang and Yuchang Su and Yiming Liu and
            Xiaohan Wang and James Burgess and Elaine Sui and
            Chenyu Wang and Josiah Aklilu and Alejandro Lozano and
            Anjiang Wei and Ludwig Schmidt and Serena Yeung-Levy},
  title  = {Automated Generation of Challenging Multiple-Choice
            Questions for Vision Language Model Evaluation},
  year   = {2025},
  url    = {https://arxiv.org/abs/2501.03225}
}

@misc{offerijns2020betterdistractionstransformerbaseddistractor,
  author = {Jeroen Offerijns and Suzan Verberne and
            Tessa Verhoef},
  title  = {Better Distractions: Transformer-based Distractor
            Generation and Multiple Choice Question Filtering},
  year   = {2020},
  url    = {https://arxiv.org/abs/2010.09598}
}

@inproceedings{papineni-etal-2002-bleu,
  address   = {Philadelphia, Pennsylvania, USA},
  author    = {Papineni, Kishore and Roukos, Salim and Ward, Todd and
               Zhu, Wei-Jing},
  booktitle = {Proceedings of the 40th Annual Meeting of the
               Association for Computational Linguistics},
  editor    = {Isabelle, Pierre and Charniak, Eugene and
               Lin, Dekang},
  month     = jul,
  pages     = {311--318},
  publisher = {Association for Computational Linguistics},
  title     = {{B}leu: a Method for Automatic Evaluation of Machine
               Translation},
  year      = {2002},
  doi       = {10.3115/1073083.1073135},
  url       = {https://aclanthology.org/P02-1040/}
}

@inproceedings{lin-2004-rouge,
  address   = {Barcelona, Spain},
  author    = {Lin, Chin-Yew},
  booktitle = {Text Summarization Branches Out},
  month     = jul,
  pages     = {74--81},
  publisher = {Association for Computational Linguistics},
  title     = {{ROUGE}: A Package for Automatic Evaluation of
               Summaries},
  year      = {2004},
  url       = {https://aclanthology.org/W04-1013/}
}

@inproceedings{liang-etal-2018-distractor,
  address   = {New Orleans, Louisiana},
  author    = {Liang, Chen and Yang, Xiao and Dave, Neisarg and
               Wham, Drew and Pursel, Bart and Giles, C. Lee},
  booktitle = {Proceedings of the Thirteenth Workshop on Innovative
               Use of {NLP} for Building Educational Applications},
  editor    = {Tetreault, Joel and Burstein, Jill and
               Kochmar, Ekaterina and Leacock, Claudia and
               Yannakoudakis, Helen},
  month     = jun,
  pages     = {284--290},
  publisher = {Association for Computational Linguistics},
  title     = {Distractor Generation for Multiple Choice Questions
               Using Learning to Rank},
  year      = {2018},
  doi       = {10.18653/v1/W18-0533},
  url       = {https://aclanthology.org/W18-0533/}
}

@inproceedings{chiang-etal-2022-cdgp,
  address   = {Abu Dhabi, United Arab Emirates},
  author    = {Chiang, Shang-Hsuan and Wang, Ssu-Cheng and
               Fan, Yao-Chung},
  booktitle = {Findings of the Association for Computational
               Linguistics: EMNLP 2022},
  editor    = {Goldberg, Yoav and Kozareva, Zornitsa and Zhang, Yue},
  month     = dec,
  pages     = {5835--5840},
  publisher = {Association for Computational Linguistics},
  title     = {{CDGP}: Automatic Cloze Distractor Generation based
               on Pre-trained Language Model},
  year      = {2022},
  doi       = {10.18653/v1/2022.findings-emnlp.429},
  url       = {https://aclanthology.org/2022.findings-emnlp.429/}
}

@inproceedings{yu-etal-2024-enhancing,
  address   = {Bangkok, Thailand},
  author    = {Yu, Han Cheng and Shih, Yu An and Law, Kin Man and
               Hsieh, KaiYu and Cheng, Yu Chen and Ho, Hsin Chih and
               Lin, Zih An and Hsu, Wen-Chuan and Fan, Yao-Chung},
  booktitle = {Findings of the Association for Computational
               Linguistics: ACL 2024},
  editor    = {Ku, Lun-Wei and Martins, Andre and Srikumar, Vivek},
  month     = aug,
  pages     = {11019--11029},
  publisher = {Association for Computational Linguistics},
  title     = {Enhancing Distractor Generation for Multiple-Choice
               Questions with Retrieval Augmented Pretraining and
               Knowledge Graph Integration},
  year      = {2024},
  doi       = {10.18653/v1/2024.findings-acl.655},
  url       = {https://aclanthology.org/2024.findings-acl.655/}
}

@inproceedings{2024.EDM-long-papers.1,
  address   = {Atlanta, Georgia, USA},
  author    = {Bilal Ghanem and Alona Fyshe},
  booktitle = {Proceedings of the 17th International Conference on
               Educational Data Mining},
  editor    = {Benjamin Paa脽en and Carrie Demmans Epp},
  month     = {July},
  pages     = {6--17},
  publisher = {International Educational Data Mining Society},
  title     = {DISTO: Textual Distractors for Multiple Choice
               Reading Comprehension Questions using Negative
               Sampling},
  year      = {2024},
  doi       = {10.5281/zenodo.12729766},
  isbn      = {978-1-7336736-5-5}
}

@inproceedings{qu-etal-2024-unsupervised,
  address   = {Bangkok, Thailand},
  author    = {Qu, Fanyi and Sun, Hao and Wu, Yunfang},
  booktitle = {Findings of the Association for Computational
               Linguistics: ACL 2024},
  editor    = {Ku, Lun-Wei and Martins, Andre and Srikumar, Vivek},
  month     = aug,
  pages     = {827--838},
  publisher = {Association for Computational Linguistics},
  title     = {Unsupervised Distractor Generation via Large Language
               Model Distilling and Counterfactual Contrastive
               Decoding},
  year      = {2024},
  doi       = {10.18653/v1/2024.findings-acl.47},
  url       = {https://aclanthology.org/2024.findings-acl.47/}
}

@misc{feng2024exploring,
  author = {Wanyong Feng and Jaewook Lee and Hunter McNichols and
            Alexander Scarlatos and Digory Smith and
            Simon Woodhead and Nancy Otero Ornelas and
            Andrew Lan},
  title  = {Exploring Automated Distractor Generation for Math
            Multiple-choice Questions via Large Language Models},
  year   = {2024},
  url    = {https://arxiv.org/abs/2404.02124}
}

@misc{scarlatos2024improving,
  author = {Alexander Scarlatos and Wanyong Feng and Digory Smith and
            Simon Woodhead and Andrew Lan},
  title  = {Improving Automated Distractor Generation for Math
            Multiple-choice Questions with Overgenerate-and-rank},
  year   = {2024},
  url    = {https://arxiv.org/abs/2405.05144}
}

@misc{fernandez2024divert,
  author = {Nigel Fernandez and Alexander Scarlatos and
            Wanyong Feng and Simon Woodhead and Andrew Lan},
  title  = {DiVERT: Distractor Generation with Variational Errors
            Represented as Text for Math Multiple-choice
            Questions},
  year   = {2024},
  url    = {https://arxiv.org/abs/2406.19356}
}

@misc{sonkar2025imitation,
  author = {Shashank Sonkar and Naiming Liu and Xinghe Chen and
            Richard G. Baraniuk},
  title  = {The Imitation Game for Educational AI},
  year   = {2025},
  url    = {https://arxiv.org/abs/2502.15127}
}

@misc{liu2025natural,
  author = {Naiming Liu and Shashank Sonkar and
            Richard G. Baraniuk},
  title  = {Do LLMs Make Mistakes Like Students? Exploring
            Natural Alignment between Language Models and Human
            Error Patterns},
  year   = {2025},
  url    = {https://arxiv.org/abs/2502.15140}
}

@misc{huang2025accuracyrobustnessstudyrule,
  author = {Yuzhen Huang and Weihao Zeng and Xingshan Zeng and
            Qi Zhu and Junxian He},
  title  = {From Accuracy to Robustness: A Study of Rule- and
            Model-based Verifiers in Mathematical Reasoning},
  year   = {2025},
  url    = {https://arxiv.org/abs/2505.22203}
}

@misc{shao2025spuriousrewardsrethinkingtraining,
  author = {Rulin Shao and Shuyue Stella Li and Rui Xin and
            Scott Geng and Yiping Wang and Sewoong Oh and
            Simon Shaolei Du and Nathan Lambert and Sewon Min and
            Ranjay Krishna and Yulia Tsvetkov and
            Hannaneh Hajishirzi and Pang Wei Koh and
            Luke Zettlemoyer},
  title  = {Spurious Rewards: Rethinking Training Signals in
            RLVR},
  year   = {2025},
  url    = {https://arxiv.org/abs/2506.10947}
}

@misc{wu2025reasoningmemorizationunreliableresults,
  author = {Mingqi Wu and Zhihao Zhang and Qiaole Dong and
            Zhiheng Xi and Jun Zhao and Senjie Jin and
            Xiaoran Fan and Yuhao Zhou and Huijie Lv and
            Ming Zhang and Yanwei Fu and Qin Liu and
            Songyang Zhang and Qi Zhang},
  title  = {Reasoning or Memorization? Unreliable Results of
            Reinforcement Learning Due to Data Contamination},
  year   = {2025},
  url    = {https://arxiv.org/abs/2507.10532}
}

@misc{wang2024mmluprorobustchallengingmultitask,
  author = {Yubo Wang and Xueguang Ma and Ge Zhang and
            Yuansheng Ni and Abhranil Chandra and Shiguang Guo and
            Weiming Ren and Aaran Arulraj and Xuan He and
            Ziyan Jiang and Tianle Li and Max Ku and Kai Wang and
            Alex Zhuang and Rongqi Fan and Xiang Yue and
            Wenhu Chen},
  title  = {MMLU-Pro: A More Robust and Challenging Multi-Task
            Language Understanding Benchmark},
  year   = {2024},
  url    = {https://arxiv.org/abs/2406.01574}
}

@misc{medqa,
  author = {Di Jin and Eileen Pan and Nassim Oufattole and
            Wei-Hung Weng and Hanyi Fang and Peter Szolovits},
  title  = {What Disease does this Patient Have? A Large-scale
            Open Domain Question Answering Dataset from Medical
            Exams},
  year   = {2020},
  url    = {https://arxiv.org/abs/2009.13081}
}

@misc{pteam2025supergpqascalingllmevaluation,
  author = {P Team and Xinrun Du and Yifan Yao and Kaijing Ma and
            Bingli Wang and Tianyu Zheng and King Zhu and
            Minghao Liu and Yiming Liang and Xiaolong Jin and
            Zhenlin Wei and Chujie Zheng and Kaixin Deng and
            Shawn Gavin and Shian Jia and Sichao Jiang and
            Yiyan Liao and Rui Li and Qinrui Li and Sirun Li and
            Yizhi Li and Yunwen Li and David Ma and Yuansheng Ni and
            Haoran Que and Qiyao Wang and Zhoufutu Wen and
            Siwei Wu and Tyshawn Hsing and Ming Xu and
            Zhenzhu Yang and Zekun Moore Wang and Junting Zhou and
            Yuelin Bai and Xingyuan Bu and Chenglin Cai and
            Liang Chen and Yifan Chen and Chengtuo Cheng and
            Tianhao Cheng and Keyi Ding and Siming Huang and
            Yun Huang and Yaoru Li and Yizhe Li and Zhaoqun Li and
            Tianhao Liang and Chengdong Lin and Hongquan Lin and
            Yinghao Ma and Tianyang Pang and Zhongyuan Peng and
            Zifan Peng and Qige Qi and Shi Qiu and Xingwei Qu and
            Shanghaoran Quan and Yizhou Tan and Zili Wang and
            Chenqing Wang and Hao Wang and Yiya Wang and
            Yubo Wang and Jiajun Xu and Kexin Yang and
            Ruibin Yuan and Yuanhao Yue and Tianyang Zhan and
            Chun Zhang and Jinyang Zhang and Xiyue Zhang and
            Xingjian Zhang and Yue Zhang and Yongchi Zhao and
            Xiangyu Zheng and Chenghua Zhong and Yang Gao and
            Zhoujun Li and Dayiheng Liu and Qian Liu and
            Tianyu Liu and Shiwen Ni and Junran Peng and
            Yujia Qin and Wenbo Su and Guoyin Wang and Shi Wang and
            Jian Yang and Min Yang and Meng Cao and Xiang Yue and
            Zhaoxiang Zhang and Wangchunshu Zhou and Jiaheng Liu and
            Qunshu Lin and Wenhao Huang and Ge Zhang},
  title  = {SuperGPQA: Scaling LLM Evaluation across 285 Graduate
            Disciplines},
  year   = {2025},
  url    = {https://arxiv.org/abs/2502.14739}
}

@misc{MedXpertQA,
  author = {Yuxin Zuo and Shang Qu and Yifei Li and Zhangren Chen and
            Xuekai Zhu and Ermo Hua and Kaiyan Zhang and
            Ning Ding and Bowen Zhou},
  title  = {MedXpertQA: Benchmarking Expert-Level Medical
            Reasoning and Understanding},
  year   = {2025},
  url    = {https://arxiv.org/abs/2501.18362}
}

@inproceedings{jin-etal-2019-pubmedqa,
  address   = {Hong Kong, China},
  author    = {Jin, Qiao and Dhingra, Bhuwan and Liu, Zhengping and
               Cohen, William and Lu, Xinghua},
  booktitle = {Proceedings of the 2019 Conference on Empirical
               Methods in Natural Language Processing and the 9th
               International Joint Conference on Natural Language
               Processing (EMNLP-IJCNLP)},
  editor    = {Inui, Kentaro and Jiang, Jing and Ng, Vincent and
               Wan, Xiaojun},
  month     = nov,
  pages     = {2567--2577},
  publisher = {Association for Computational Linguistics},
  title     = {{P}ub{M}ed{QA}: A Dataset for Biomedical Research
               Question Answering},
  year      = {2019},
  doi       = {10.18653/v1/D19-1259},
  url       = {https://aclanthology.org/D19-1259/}
}

@inproceedings{kool2019buy,
  author    = {Wouter Kool and Herke van Hoof and Max Welling},
  booktitle = {DeepRLStructPred@ICLR},
  title     = {Buy 4 REINFORCE Samples, Get a Baseline for Free!},
  year      = {2019},
  url       = {https://api.semanticscholar.org/CorpusID:198489118}
}

\newpage
\appendix

\section{Experimental Hyperparameter Settings}
\label{sec:hyperparameters}

All experiments use the same hyperparameter configuration to ensure comparability.
The base models are Llama-3.1-8B-Instruct and Qwen2-7B-Instruct, trained using the GRPO algorithm, with RLOO~\citep{kool2019buy} employed as the advantage estimator.
All experiments are conducted on a single node equipped with 8 H200 GPUs.

\noindent\textbf{Batch Size and Sampling.} The training batch size is 128, and the mini-batch size is also 128 (i.e., fully on-policy).
We sample 8 responses per question for group-wise advantage estimation.

\noindent\textbf{Optimizer Configuration.} The learning rate is set to 1e-6 using a cosine warmup strategy. The warmup duration is 10 steps, and the minimum learning rate is 0.1 times the initial learning rate.
The weight decay coefficient is 0.1, and the gradient clipping threshold is 1.0.

\noindent\textbf{Reinforcement Learning Settings.} We use KL divergence loss to constrain policy updates, with a coefficient of 0.001.
The entropy regularization coefficient is 0.001.

\noindent\textbf{Sequence Length.} Both the maximum prompt length and maximum response length are set to 2048 tokens.

\noindent\textbf{Generation Configuration.} For training, sampling uses temperature=1.0, top-p=1.0, and top-k=$-1$; for testing, we use greedy decoding.

\section{MMLU-Pro Data Preprocessing Details}
\label{sec:data-preprocessing}

This section details the preprocessing procedure for the MMLU-Pro dataset used in Section~\ref{sec:number_distractor}.

The original MMLU-Pro~\citep{wang2024mmluprorobustchallengingmultitask} test set comprises 12,032 questions, approximately 83\% of which are 10-choice questions, with the remainder being 3-9 choice questions.
We retain only the 10-choice questions to ensure consistency in subsequent processing. We remove a small number of duplicate samples (based on exact matches of question stems and options), resulting in 9,417 10-choice samples for experiments. These are randomly split into a training set (8,004 samples) and a test set (1,413 samples), maintaining an approximate ratio of 85:15.

\section{Detailed Cross-Option Generalization Results}
\label{sec:full_results_appendix}

This section presents the full experimental results of cross-option generalization, corresponding to the analysis in Section~\ref{sec:number_distractor}.

\begin{table*}[ht]
\centering
\small
\begin{minipage}{0.48\textwidth}
\centering
\begin{tabular}{l|ccccc|c}
\toprule
\multirow{2}{*}{\textbf{Train}} & \multicolumn{5}{c|}{\textbf{Test}} & \multirow{2}{*}{\textbf{Avg}} \\
\cmidrule(lr){2-6}
& \textbf{2} & \textbf{4} & \textbf{6} & \textbf{8} & \textbf{10} & \\
\midrule
\multicolumn{1}{c|}{\textbf{2}} & \cellcolor{orange!49}\textbf{80.27} & \cellcolor{orange!5}67.82 & \cellcolor{orange!5}60.50 & \cellcolor{orange!5}54.28 & \cellcolor{orange!5}52.44 & 63.06 \\
\multicolumn{1}{c|}{\textbf{4}} & \cellcolor{orange!45}\underline{79.95} & \cellcolor{orange!44}\textbf{69.96} & \cellcolor{orange!35}63.79 & \cellcolor{orange!27}57.40 & \cellcolor{orange!29}56.19 & 65.46 \\
\multicolumn{1}{c|}{\textbf{6}} & \cellcolor{orange!40}79.57 & \cellcolor{orange!35}69.58 & \cellcolor{orange!42}\textbf{64.07} & \cellcolor{orange!38}59.21 & \cellcolor{orange!39}57.87 & \textbf{66.06} \\
\multicolumn{1}{c|}{\textbf{8}} & \cellcolor{orange!9}76.76 & \cellcolor{orange!38}69.65 & \cellcolor{orange!40}\underline{63.93} & \cellcolor{orange!47}\textbf{60.92} & \cellcolor{orange!40}\underline{58.17} & \underline{65.89} \\
\multicolumn{1}{c|}{\textbf{10}} & \cellcolor{orange!11}76.98 & \cellcolor{orange!41}\underline{69.79} & \cellcolor{orange!34}63.32 & \cellcolor{orange!42}\underline{60.19} & \cellcolor{orange!42}\textbf{58.61} & 65.78 \\
\midrule
\textbf{Avg} & 78.71 & 69.36 & 63.12 & 58.40 & 56.66 & 65.25 \\
\bottomrule
\end{tabular}
\subcaption{Llama-3.1-8B}
\label{tab:full_results_llama}
\end{minipage}
\hfill
\begin{minipage}{0.48\textwidth}
\centering
\begin{tabular}{l|ccccc|c}
\toprule
\multirow{2}{*}{\textbf{Train}} & \multicolumn{5}{c|}{\textbf{Test}} & \multirow{2}{*}{\textbf{Avg}} \\
\cmidrule(lr){2-6}
& \textbf{2} & \textbf{4} & \textbf{6} & \textbf{8} & \textbf{10} & \\
\midrule
\multicolumn{1}{c|}{\textbf{2}} & \cellcolor{orange!55}\textbf{77.61} & \cellcolor{orange!35}64.19 & \cellcolor{orange!10}56.34 & \cellcolor{orange!5}49.50 & \cellcolor{orange!5}46.66 & 58.86 \\
\multicolumn{1}{c|}{\textbf{4}} & \cellcolor{orange!33}\underline{72.15} & \cellcolor{orange!48}\textbf{65.15} & \cellcolor{orange!51}\textbf{58.22} & \cellcolor{orange!29}51.92 & \cellcolor{orange!28}48.57 & \textbf{59.20} \\
\multicolumn{1}{c|}{\textbf{6}} & \cellcolor{orange!28}71.17 & \cellcolor{orange!37}\underline{64.36} & \cellcolor{orange!35}57.49 & \cellcolor{orange!39}\underline{52.71} & \cellcolor{orange!44}\underline{49.68} & \underline{59.08} \\
\multicolumn{1}{c|}{\textbf{8}} & \cellcolor{orange!11}67.68 & \cellcolor{orange!5}61.59 & \cellcolor{orange!10}56.34 & \cellcolor{orange!32}52.13 & \cellcolor{orange!25}48.37 & 57.22 \\
\multicolumn{1}{c|}{\textbf{10}} & \cellcolor{orange!16}68.68 & \cellcolor{orange!28}63.67 & \cellcolor{orange!42}\underline{57.82} & \cellcolor{orange!49}\textbf{53.50} & \cellcolor{orange!49}\textbf{50.07} & 58.75 \\
\midrule
\textbf{Avg} & 71.46 & 63.79 & 57.24 & 51.95 & 48.67 & 58.62 \\
\bottomrule
\end{tabular}
\subcaption{Qwen2-7B}
\label{tab:full_results_qwen}
\end{minipage}
\caption{Cross-option generalization results (Accuracy \%). \textbf{Bold} and \underline{Underlined} denote the best and second-best results in each column (excluding the mix strategy). Background color intensity represents normalized score $z$, with darker colors indicating better performance.}
\label{tab:full_results}
\end{table*}

Table~\ref{tab:full_results} reveals a clear diagonal pattern: the highest accuracies and darkest shading concentrate along the diagonal where training and test option counts match.
For Llama-3.1, the best performance consistently appears on the diagonal. 
Qwen2 shows minor deviations (e.g., 4-train peaks on 6-test, 10-train on 8-test), but generally confirms that proximity to the test option-count matters more than the absolute count.
This suggests that the number of training options has a limited impact on performance compared to the alignment between training and test option counts.

\section{Experimental Setup for Option Expansion}
\label{sec:experimental_setup_for_option_expansion}

\subsection{Detailed Experimental Configuration}
\label{subsec:detail_experimental_configurations_for_table2}

This section provides the detailed experimental configuration for the option expansion experiments reported in Table~\ref{tab:option_expansion_results}.

\noindent\textbf{Models.} We use three types of option generator models for distractor synthesis, all with temperature set to 0.7:
\begin{itemize}
    \item \textbf{GPT-OSS-120B}: The GPT-OSS-120B model with high reasoning effort enabled.
    \item \textbf{Qwen2.5-32B-Instruct}: The Qwen2.5-32B-Instruct model.
    \item \textbf{Self}: The target model itself (Qwen2-7B or Llama-3.1-8B) is used to generate distractors.
\end{itemize}

\noindent\textbf{Output Postprocessing.} Model outputs are postprocessed using a regex-based cleaning script to remove common output prefixes. Specifically, we strip patterns such as \texttt{"Option A: xxxxx"}, \texttt{"I. xxxxx"}, or similar formatting, retaining only the core option content itself. This ensures that the generated distractors are clean and consistent with the original option format.

\subsection{Option Expansion Prompt}
\label{subsec:option-expansion-prompt}

This section provides the full prompt used for option expansion experiments in Section~\ref{sec:number_distractor}.

\definecolor{deepseekblue}{RGB}{0,102,204}

\begin{tcolorbox}[
  enhanced,
  breakable,
  colback=white, 
  colframe=deepseekblue,  
  width=\columnwidth,
  arc=2mm, 
  boxrule=0.5mm, 
  title={\normalsize\textbf{Prompt:} Option Expansion},
  fonttitle=\bfseries\normalsize, 
  fontupper=\footnotesize,
]
\begin{verbatim}
I have a multiple-choice question with 
{current_num} options, one of which is 
correct, and I need to expand it to a 
{target_num}-option multiple-choice question.
Please generate {new_options_num} additional 
plausible but incorrect options to accompany 
the original options.

Requirements:
1. The new options should be plausible and 
   similar in style to the existing options
2. The new options should be clearly incorrect 
   (not ambiguous)
3. The new options should test different 
   aspects of the question topic
4. Please think step by step about how to 
   create good distractors

IMPORTANT - Output Format:
You MUST output ONLY a valid JSON object 
wrapped in ```json and ``` markers.
Do NOT include any other text before or after 
the JSON code block.
The JSON must have exactly these two fields:
- "thinking": a string containing your 
  reasoning process
- "new_options": an array of exactly 
  {new_options_num} strings 

Example format:
```json
{
  "thinking": "I will create options that are 
    similar to the existing ones but test 
    different concepts...",
  "new_options": ["Option 1", "Option 2", 
    "Option 3"]
}
```

Here is the original question:
<original question>
{raw_question}
</original question>

Now generate exactly {new_options_num} new 
incorrect options in the JSON format specified.
\end{verbatim}
\end{tcolorbox}

\section{Baseline Prompts}
\label{sec:baseline_prompts}

This section provides the detailed prompts used in the baseline methods described in Section~\ref{sec:experimental_setup}.

\subsection{Filtering Prompt for Evaluating Question Convertibility}

The following prompt is used to evaluate whether a multiple-choice question can be transformed into a standalone QA pair:

\begin{tcolorbox}[
  enhanced,
  breakable,
  colback=white, 
  colframe=deepseekblue,  
  width=\columnwidth,
  arc=2mm, 
  boxrule=0.5mm, 
  title={\normalsize\textbf{Prompt:} Question Convertibility Filter},
  fonttitle=\bfseries\normalsize, 
  fontupper=\footnotesize,
]
\begin{verbatim}
You are an expert in educational assessment. 
Evaluate whether the following Multiple-Choice 
Question can be transformed into a standalone 
QA pair.

Original Question: {question}

Correct Answer: {correct_answer}

Task:
Determine if the stem is understandable 
and answerable without seeing the options.

Criteria for "NOT_CONVERTIBLE":
- Contains phrases like "Which of the 
  following", "EXCEPT", "NOT true", 
  "All of the above".
- Requires comparing options to find the "best" 
  answer (e.g., "best describes", "best 
  explains").
- The answer depends entirely on the specific 
  choices given.
- Refers to figures, graphs, or tables that are 
  part of the options.

Criteria for "CONVERTIBLE":
- The question has a clear, specific answer 
  that doesn't depend on seeing options.
- The question is complete and self-contained.

CRITICAL: You MUST end your response with one 
of these EXACT lines:
FINAL_LABEL: CONVERTIBLE
OR
FINAL_LABEL: NOT_CONVERTIBLE

Example Output 1:
Analysis: The question asks "Which of the 
following is a fruit?" which implies a 
selection from a specific list. Without
options, the answer is too open-ended.
FINAL_LABEL: NOT_CONVERTIBLE

Example Output 2:
Analysis: The question asks "What is the most 
likely diagnosis?" which is clear and specific. 
It can be answered without seeing the options.
FINAL_LABEL: CONVERTIBLE

Your Output:
\end{verbatim}
\end{tcolorbox}

\subsection{Model-Based Conversion Prompt}

The following prompt is used to convert multiple-choice questions to fill-in-the-blank or short-answer format:

\begin{tcolorbox}[
  enhanced,
  breakable,
  colback=white, 
  colframe=deepseekblue,  
  width=\columnwidth,
  arc=2mm, 
  boxrule=0.5mm, 
  title={\normalsize\textbf{Prompt:} Model-Based Conversion},
  fonttitle=\bfseries\normalsize, 
  fontupper=\footnotesize,
]
\begin{verbatim}
I will give you a multiple-choice question with 
answer options and its correct answer. Please 
convert it into a fill-in-the-blank or 
short-answer question.

IMPORTANT INSTRUCTIONS:
1. PRESERVE ALL CONTEXT: Keep the entire 
   clinical scenario, patient information, 
   symptoms, lab values, physical exam 
   findings, and background information 
   EXACTLY as given. Do NOT summarize, 
   shorten, or remove any details.

2. ONLY modify the question format: Remove the 
   answer options (A, B, C, D) and convert the 
   final question into a fill-in-the-blank or 
   short-answer format.

3. The modified question should ask for the
   same information, but without providing the 
   multiple-choice options.

4. Provide the correct answer as a short, 
   direct response (no explanation needed).

Original question with options:
{question}

Correct answer:
{answer}

Return the output using this format:
<Question>{Modified Question - with ALL 
original context preserved}</Question>
<Answer>{Short Answer}</Answer>

Example:
If given: "A 50-year-old man presents with 
chest pain... BP 120/80... What is the 
diagnosis? A. MI B. Angina C. PE D. GERD"

You should return: "A 50-year-old man presents 
with chest pain... BP 120/80... What is the 
diagnosis?" (keeping all context, just 
removing A/B/C/D options)
\end{verbatim}
\end{tcolorbox}

\section{Experimental Setup for Distractor Strength Analysis}
\label{sec:distractor_strength_setup}

This section provides the detailed experimental configuration for the distractor strength analysis experiments reported in Section~\ref{sec:distractor_strength_analysis} and Table~\ref{tab:distractor_strength_results}.
To estimate the empirical strength $\hat{s}_j$ of each distractor $o_j$, we adopt the following procedure:

\noindent\textbf{Sampling Configuration.} For each question, we sample $K = 8$ trajectories from the evaluation model with temperature set to 0.7. The model generates complete outputs $\{y_k = (c_k, a_k)\}_{k=1}^{K}$, where $c_k$ is the reasoning chain and $a_k$ is the selected option.

\noindent\textbf{Distractor Selection.} After computing empirical strengths for all distractors in each question, we rank them by $\hat{s}_j$ in descending order. For the experiments in Table~\ref{tab:distractor_strength_results}:
\begin{itemize}
    \item \textbf{mcq2-random}: Randomly select one distractor from the original 9 distractors.
    \item \textbf{w/o [Model]}: Retain the distractor with the lowest empirical strength estimated by the specified model.
    \item \textbf{w/ [Model]}: Retain the distractor with the highest empirical strength estimated by the specified model.
\end{itemize}

\section{Experimental Setup for Rejection Sampling Analysis}
\label{sec:rs_setup}

This section details the experimental setup for the rejection sampling analysis presented in Figure~\ref{fig:rs_rounds}.
The y-axis in Figure~\ref{fig:rs_rounds} represents the mean accuracy of Qwen2-7B-Instruct across four out-of-distribution (OOD) datasets: SuperGPQA (Clinical), MMLU-Pro (Health), MedXpertQA, and PubMedQA.
For the single-round baseline in Figure~\ref{fig:rs_rounds}, we use Qwen2.5-32B-Instruct as the generator model and do not use a strength estimator; the model directly rewrites all distractors into a new option set, as shown in the prompt below.

\begin{tcolorbox}[
  enhanced,
  breakable,
  colback=white, 
  colframe=deepseekblue,  
  width=\columnwidth,
  arc=2mm, 
  boxrule=0.5mm, 
  title={\normalsize\textbf{Prompt:} Direct Distractor Rewriting},
  fonttitle=\bfseries\normalsize, 
  fontupper=\footnotesize,
]
\begin{verbatim}
You are a professional expert in generating 
distractors for multiple-choice questions. 
Your task is to evaluate the existing 
distractors and improve them if necessary.

Question Information:
Question Text: {question_text}
Correct Option: {correct_option}
Correct Answer: {correct_answer}
Existing Options (with all options 
including correct answer): {all_options}

Your Task:
Review the existing distractors (incorrect 
options). You have TWO choices:
1. If you think the existing distractors are 
   already reasonable and effective, you 
   can KEEP them as-is
2. If you think some or all distractors need 
   improvement, REPLACE only those that need 
   improvement

Requirements for generating/keeping 
distractors:
1. Distractors must be plausible and able to 
   attract students who are not careful enough 
   or have insufficient knowledge
2. Distractors should be consistent with the 
   correct answer in form and style
3. **CRITICAL**: Distractors MUST NOT be the 
   same as or similar to the correct answer 
   "{correct_answer}"
4. **CRITICAL**: Do NOT use the correct option 
   identifier "{correct_option}" as a key in 
   your output
5. **CRITICAL**: ONLY use the SAME option 
   identifiers as the existing distractors. For 
   example, if existing distractors use keys A, 
   C, D, your output must ONLY use keys A, C, D
6. Distractors should have discrimination, 
   avoiding duplication or being too similar to 
   each other
7. If you decide to keep existing distractors, 
   output them exactly as they are

Decision:
First, decide whether you want to KEEP all 
existing distractors or IMPROVE some/all of 
them.

Output Format:
Please return the result in JSON format with 
the following fields:
- decision: string type, either "KEEP_ALL" or 
  "IMPROVE"
- distractors: dictionary type, with keys as 
  option identifiers (MUST use the same keys as 
  existing distractors)
  - If decision is "KEEP_ALL", output the 
    original distractors unchanged
  - If decision is "IMPROVE", output the 
    improved distractors (you can keep some 
    and improve others)
- reasoning: string type, explaining your 
  decision and rationale

Example 1 (Keep all):
If existing options are: {"A": "distractor 1", 
"B": "correct answer", "C": "distractor 2", 
"D": "distractor 3"}
And correct option is B, you might output:
{"decision": "KEEP_ALL", "distractors": 
{"A": "distractor 1", "C": "distractor 2", 
"D": "distractor 3"}, "reasoning": "The 
existing distractors are already well-designed 
and plausible."}

Example 2 (Improve some):
If existing options are: {"A": "obviously 
wrong", "B": "correct answer", "C": 
"reasonable distractor", "D": "too similar 
to B"}
And correct option is B, you might output:
{"decision": "IMPROVE", "distractors": 
{"A": "improved distractor 1", "C": 
"reasonable distractor", "D": "improved 
distractor 2"}, 
"reasoning": "Options A and D need 
improvement, but C is already good."}
\end{verbatim}
\end{tcolorbox}

\section{Prompts for Iterative Distractor Curation}
\label{sec:idc_prompts}

This section provides the detailed prompts used in the Iterative Distractor Curation framework described in Section~\ref{sec:idc}.
The framework employs three key prompts to ensure the quality and effectiveness of generated distractors: (1) a prompt for generating new distractor candidates, (2) a prompt for evaluating model performance on the multiple-choice question, and (3) a prompt for semantic equivalence checking to ensure generated distractors do not become alternative correct answers.

\subsection{Distractor Generation Prompt}

The following prompt is used to generate new distractor candidates during both the filling and replacement phases:

\begin{tcolorbox}[
  enhanced,
  breakable,
  colback=white, 
  colframe=deepseekblue,  
  width=\columnwidth,
  arc=2mm, 
  boxrule=0.5mm, 
  title={\normalsize\textbf{Prompt:} Distractor Generation},
  fonttitle=\bfseries\normalsize, 
  fontupper=\footnotesize,
]
\begin{verbatim}
You are a professional expert in generating 
distractors for multiple-choice questions. 
Your task is to generate high-quality 
REPLACEMENT distractors for the given 
question.

Question Information:
Question Text: {question_text}
Correct Option: {correct_option}
Correct Answer: {correct_answer}
Number of Distractors to Generate: 
{num_distractors} 
Existing Distractors (to be replaced): 
{existing_distractors}

Your Task:
You need to REPLACE the existing distractors 
with better, more effective ones. The new 
distractors should be more challenging and 
attractive to students who lack careful 
thinking or sufficient knowledge.

Requirements:
1. The generated distractors must be plausible 
   and able to attract students who are not 
   careful enough or have insufficient 
   knowledge
2. Distractors should be consistent with the 
   correct answer in form and style
3. **CRITICAL**: Distractors MUST NOT be the 
   same as or similar to the correct answer 
   "{correct_answer}". They should be different 
   but plausible alternatives.
4. **CRITICAL**: Do NOT use the correct option 
   identifier "{correct_option}" as a key 
   in your output.
5. **CRITICAL**: ONLY use the SAME option 
   identifiers as the existing distractors. For 
   example, if existing distractors use keys 
   A, C, D, your output must ONLY use keys A, 
   C, D. Do NOT create new keys like E, F, G, etc.
6. Distractors should have discrimination, 
   avoiding duplication or being too similar 
   to each other
7. The new distractors should be more effective 
   than the existing ones - they should be 
   harder to distinguish from the correct 
   answer or more tempting to select

{failed_feedback}

Output Format:
Please return the result in JSON format with 
the following fields:
- distractors: dictionary type, with keys as 
  option identifiers (MUST use the same keys 
  as existing distractors, e.g., if existing 
  has A, C, D, output must have A, C, D)
- reasoning: string type, explaining your 
  rationale for generating these distractors

Example:
If existing distractors are: {"A": "old 
distractor 1", "C": "old distractor 2", 
"D": "old distractor 3"}
Your output should be: {"A": "new 
distractor 1", "C": "new distractor 2", 
"D": "new distractor 3"}
Do NOT output: {"E": "...", "F": "...", 
"G": "..."}
\end{verbatim}
\end{tcolorbox}

\subsection{Model Evaluation Prompt}

The following prompt is used to evaluate whether the base model can correctly answer the multiple-choice question with the generated distractors:

\begin{tcolorbox}[
  enhanced,
  breakable,
  colback=white, 
  colframe=deepseekblue,  
  width=\columnwidth,
  arc=2mm, 
  boxrule=0.5mm, 
  title={\normalsize\textbf{Prompt:} Model Evaluation},
  fonttitle=\bfseries\normalsize, 
  fontupper=\footnotesize,
]
\begin{verbatim}
You are a student taking an exam. Please read 
the following question carefully and choose 
the answer you believe is correct.

Question:
{question_text}

Options:
{options}
\end{verbatim}
\end{tcolorbox}

\subsection{Semantic Equivalence Check Prompt}

The following prompt is used to ensure that generated distractors are not semantically equivalent to the correct answer, preventing the creation of questions with multiple valid answers:

\begin{tcolorbox}[
  enhanced,
  breakable,
  colback=white, 
  colframe=deepseekblue,  
  width=\columnwidth,
  arc=2mm, 
  boxrule=0.5mm, 
  title={\normalsize\textbf{Prompt:} Semantic Equivalence Check},
  fonttitle=\bfseries\normalsize, 
  fontupper=\footnotesize,
]
\begin{verbatim}
You are an expert in evaluating whether two 
answer options in a multiple-choice question 
are semantically equivalent. Your task is to 
determine if Text 2 conveys EXACTLY the same 
meaning as Text 1 (the correct answer), such 
that both would be equally correct.

Question Context:
{question_text}

Text 1 (Correct Answer): {text1}
Text 2 (Generated Distractor - to be checked): 
{text2}

CRITICAL: Are these two texts IDENTICAL in 
meaning? Would both be equally correct as 
the answer to the question above?

Evaluation Guidelines:
1. **Be VERY STRICT**: Only mark as EQUIVALENT 
   if the texts are COMPLETELY identical in 
   meaning
2. **Context matters**: Evaluate the texts 
   specifically in the context of the question 
   above

Please respond with ONLY ONE of the following:
- "EQUIVALENT" - if and only if the texts are 
  completely identical in meaning and would 
  both be equally correct
- "NOT_EQUIVALENT" - if there is ANY difference 
  in meaning, specificity, certainty, or scope

Your response:
\end{verbatim}
\end{tcolorbox}

\section{Case Study: Iterative Distractor Evolution}
\label{sec:case_study}

This section provides a detailed case study illustrating how the Iterative Distractor Curation framework progressively strengthens distractors through the filling and replacement phases.
We trace the evolution of a medical question from MedQA that initially had a passrate of 1.0 (the model was not distracted by any incorrect options).

\subsection{Original Question}

The original question contains weak distractors that fail to challenge the model:

\begin{tcolorbox}[
  enhanced,
  breakable,
  colback=white, 
  colframe=deepseekblue,  
  width=\columnwidth,
  arc=2mm, 
  boxrule=0.5mm, 
  title={\normalsize\textbf{Initial State:} Passrate = 1.0},
  fonttitle=\bfseries\normalsize, 
  fontupper=\footnotesize,
]
\begin{verbatim}
Question: A 58-year-old man comes to the 
emergency department for complaints of 
crushing chest pain for 4 hours. He was 
shoveling snow outside when the pain started. 
It is rated 7/10 and radiates to his left arm. 
An electrocardiogram (ECG) demonstrates 
ST-segment elevation in leads V2-4. He 
subsequently undergoes percutaneous coronary 
intervention and is discharged with aspirin, 
clopidogrel, carvedilol, atorvastatin, and 
lisinopril. Five days later, the patient is 
brought to the emergency department by his wife 
with complaints of dizziness. He reports 
lightheadedness and palpitations for the past 
2 hours but otherwise feels fine. His 
temperature is 99.7°F (37.6°C), blood pressure 
is 95/55 mmHg, pulse is 105/min, and 
respirations are 17/min. A pulmonary artery 
catheter is performed and demonstrates an 
increase in oxygen concentration at the 
pulmonary artery. What finding would you 
expect in this patient?

A. Midsystolic click followed by a late 
   systolic murmur at the apex
B. Low-pitched, rumbling diastolic murmur at 
   the mid-left sternal border
C. Soft, blowing systolic murmur heard best at 
   the apex
D. Harsh, loud, holosystolic murmur at the 
   lower left sternal border [CORRECT]
\end{verbatim}
\end{tcolorbox}

\subsection{Filling Phase}

The algorithm begins by generating new distractors to replace the ineffective ones.

\noindent\textbf{First Successful Round:} The generator produces two effective distractors.

\begin{tcolorbox}[
  enhanced,
  breakable,
  colback=white, 
  colframe=deepseekblue,  
  width=\columnwidth,
  arc=2mm, 
  boxrule=0.5mm, 
  title={\normalsize\textbf{Filling Round 1}},
  fonttitle=\bfseries\normalsize, 
  fontupper=\footnotesize,
]
\begin{verbatim}
Generated Distractors:
- Soft, blowing diastolic murmur at the apex
- High-pitched, crescendo-decrescendo sys-
  tolic murmur at the right upper sternal border
- Continuous machinery-like murmur at 
  the left upper sternal border

Reasoning: The distractors have been chosen to 
be plausible findings that could be observed in 
a patient with similar symptoms, but each 
represents a different type of murmur or heart 
sound. The distractors are consistent in form 
and style with the correct answer, presenting 
descriptions of heart murmurs with their 
location and characteristics. Each distractor 
is chosen to represent a different cardiac 
condition, such as aortic stenosis, aortic 
regurgitation, or patent ductus arteriosus
, making them plausible but distinct 
alternatives to the correct answer.

Validation Result: The first two distractors
successfully confuse the model.
Status: 2 out of 3 distractor slots filled.
\end{verbatim}
\end{tcolorbox}

\noindent\textbf{Second Successful Round:} The generator produces one more distractor to complete the filling phase.

\begin{tcolorbox}[
  enhanced,
  breakable,
  colback=white, 
  colframe=deepseekblue,  
  width=\columnwidth,
  arc=2mm, 
  boxrule=0.5mm, 
  title={\normalsize\textbf{Filling Round 2}},
  fonttitle=\bfseries\normalsize, 
  fontupper=\footnotesize,
]
\begin{verbatim}
Generated Distractor:
-  Low-pitched, rumbling diastolic murmur at 
   the mid-left sternal border

Reasoning: The generated distractor 'Low-
pitched, rumbling diastolic murmur at the 
mid-left sternal border' is plausible and 
consistent with the form and style of the 
correct answer. It offers a different type 
of murmur (diastolic vs. systolic) and 
origin, which could distract students who 
may not fully understand the clinical context. 
This distractor is designed to challenge 
students' knowledge of cardiac murmurs and 
their ability to correlate symptoms with 
specific cardiac conditions, making it more 
effective than the existing distractors.

Validation Result: The new distractor 
successfully confuses the model.
Status: Filling phase complete 
(3/3 slots filled).
\end{verbatim}
\end{tcolorbox}

\subsection{Replacement Phase}

After the filling phase, the algorithm enters the replacement phase, where it attempts to replace weaker distractors with stronger ones.

\noindent\textbf{First Successful Round:} The generator attempts to replace the weakest distractor.

\begin{tcolorbox}[
  enhanced,
  breakable,
  colback=white, 
  colframe=deepseekblue,  
  width=\columnwidth,
  arc=2mm, 
  boxrule=0.5mm, 
  title={\normalsize\textbf{Replacement Round 1}},
  fonttitle=\bfseries\normalsize, 
  fontupper=\footnotesize,
]
\begin{verbatim}
Generated Distractor:
B. Midsystolic click followed by a late 
   systolic murmur at the apex

Reasoning: The generated distractor, 
'Midsystolic click followed by a late 
systolic murmur at the apex,' is plausible 
because it represents a common finding in 
patients with mitral valve prolapse, which 
is a condition that could present with 
symptoms of dizziness and palpitations. 
However, it is distinctly different from 
the correct answer, a harsh, loud, holosys-
tolic murmur at the lower left sternal border, 
which is indicative of acute mitral regurgi-
tation. This distractor maintains a consistent 
style with the correct answer and existing 
distractors, making it plausible yet incorrect, 
thus serving as an effective distractor.

Replacement Details:
- Replaced: "Soft, blowing diastolic murmur at 
  the apex"
- New: "Midsystolic click followed by a late 
  systolic murmur at the apex"
\end{verbatim}
\end{tcolorbox}

\begin{table*}[t]
\centering
\small
\resizebox{0.8\textwidth}{!}{
\begin{tabular}{l|cc|ccccc}
\toprule
\textbf{Dataset} & \textbf{Base} & \textbf{w/ Origin} & \textbf{Self} & \textbf{DS-V3.1} & \textbf{OSS} & \textbf{Qwen2.5-32B} & \textbf{Llama-70B} \\
\midrule
MedQA & 69.76 & 72.74 & 73.92 & 74.71 & \textbf{74.86} & 73.13 & 74.23 \\
SuperGPQA & 37.19 & 38.83 & 40.07 & 39.16 & 39.66 & 38.92 & \textbf{40.48} \\
MMLU-Pro & 63.94 & 64.79 & 65.53 & 65.16 & \textbf{65.77} & 64.55 & 64.30 \\
MedXpertQA & 14.41 & 16.41 & 18.20 & 18.04 & \textbf{18.69} & 18.65 & 17.96 \\
PubMedQA & 77.90 & 76.80 & 78.00 & 78.20 & 77.30 & 78.50 & \textbf{78.60} \\
\midrule
\textbf{Average} & 52.64 & 53.91 & 55.14 & 55.05 & \textbf{55.26} & 54.75 & 55.11 \\
\bottomrule
\end{tabular}
}
\caption{Ablation study on distractor generators for Qwen2.5-14B-Instruct. The \textbf{Base} column shows the performance of the original instruct model without RLVR training. The \textbf{w/ Origin} column shows training with original distractors from the dataset. OSS: GPT-OSS-120B; DS-V3.1: DeepSeek-V3.1.}
\label{tab:qwen14b_generator_ablation}
\end{table*}

\noindent\textbf{Second Successful Round:} The generator attempts to replace another weak distractor.

\begin{tcolorbox}[
  enhanced,
  breakable,
  colback=white, 
  colframe=deepseekblue,  
  width=\columnwidth,
  arc=2mm, 
  boxrule=0.5mm, 
  title={\normalsize\textbf{Replacement Round 2}},
  fonttitle=\bfseries\normalsize, 
  fontupper=\footnotesize,
]
\begin{verbatim}
Generated Distractor:
-  Soft, blowing systolic murmur heard best at 
   the apex

Reasoning: The generated distractor 'Soft, 
blowing systolic murmur heard best at the apex' 
is plausible and consistent with the correct 
answer in form and style. It describes a 
different type of murmur with a different 
location, which makes it a plausible but 
incorrect option for students who might not 
fully understand the specific characteristics 
of the murmur associated with the patient's 
condition. This distractor avoids duplicating 
the style or specifics of the correct answer, 
such as the harshness, loudness, and 
holosystolic nature of the murmur at the lower 
left sternal border. The distractor is 
designed to be sufficiently different from 
the correct answer and other existing 
distractors to challenge the students' 
understanding without being too obvious.

Replacement Details:
- Replaced: "High-pitched, crescendo-
  decrescendo systolic murmur at the right 
  upper sternal border"
- New: "Soft, blowing systolic murmur heard 
  best at the apex"
\end{verbatim}
\end{tcolorbox}

\subsection{Final Question State}

After completing the iterative curation process, the question now contains significantly stronger distractors:

\begin{tcolorbox}[
  enhanced,
  breakable,
  colback=white, 
  colframe=deepseekblue,  
  width=\columnwidth,
  arc=2mm, 
  boxrule=0.5mm, 
  title={\normalsize\textbf{Final State:} Enhanced Distractors},
  fonttitle=\bfseries\normalsize, 
  fontupper=\footnotesize,
]
\begin{verbatim}
Final Options:
A. Midsystolic click followed by a late 
   systolic murmur at the apex [NEW - Effective]
B. Low-pitched, rumbling diastolic murmur at 
   the mid-left sternal border [NEW - Effective]
C. Soft, blowing systolic murmur heard best at 
   the apex [NEW - Effective]
D. Harsh, loud, holosystolic murmur at the 
   lower left sternal border [CORRECT]

Summary:
- Original passrate: 1.0 (model not challenged)
- Final state: 3 effective distractors added
- The model now faces significantly higher 
  difficulty, providing clearer reward signals 
  for RLVR training
\end{verbatim}
\end{tcolorbox}

This case study demonstrates how the iterative curation framework systematically strengthens distractors through targeted generation and replacement, transforming a trivial question into a challenging learning signal for RLVR.

\section{Generalizability to Larger Models}
\label{sec:qwen14b_ablation}

To further validate the generalizability of our findings across different model scales, we conducted additional experiments using Qwen2.5-14B-Instruct as the base model.
Table~\ref{tab:qwen14b_generator_ablation} presents the results of comparing different distractor generators for this larger model.
The findings are consistent with the observations from Qwen2-7B and Llama-3.1-8B: self-generated distractors remain highly competitive.
This confirms that the effectiveness of self-generated distractors is not limited to smaller models but extends to intermediate-scale models.

\section{Impact of Estimator Selection}
\label{sec:estimator_ablation}

In our main method, we use the base model itself to estimate the empirical strength of distractors (Self-Estimation).
To validate this design choice, we conduct an ablation study where we fix the generator model (Qwen2.5-32B-Instruct) but vary the model used for strength estimation (the ``Estimator'').
We compare two settings:
(1) \textbf{Self-Estimation}: The model being trained is also used to estimate distractor strength (our default setting).
(2) \textbf{Cross-Estimation}: A different model is used to estimate distractor strength (e.g., using Llama-3.1-8B to filter distractors for Qwen2-7B).

Table~\ref{tab:estimator_ablation} presents the results.
For Qwen2-7B, Self-Estimation achieves slightly better performance across most datasets.
For Llama-3.1-8B, the results are mixed, with Cross-Estimation showing marginal gains on some tasks while Self-Estimation works better on others.
Given that self-estimation eliminates the dependency on external models and simplifies the pipeline, we conclude that it serves as a sufficient and effective strategy for filtering distractors.

\begin{table}[ht]
\centering
\small
\resizebox{\columnwidth}{!}{
\begin{tabular}{l|cc|cc}
\toprule
\multirow{2}{*}{\textbf{Dataset}} & \multicolumn{2}{c|}{\textbf{Train Qwen2-7B}} & \multicolumn{2}{c}{\textbf{Train Llama-3.1-8B}} \\
\cmidrule(lr){2-3} \cmidrule(lr){4-5}
& \textbf{Est: Self} & \textbf{Est: Llama} & \textbf{Est: Self} & \textbf{Est: Qwen} \\
\midrule
MedQA & \textbf{49.80} & 49.41 & 66.14 & \textbf{67.71} \\
SuperGPQA & \textbf{27.91} & 27.42 & \textbf{35.71} & 35.63 \\
MMLU-Pro & \textbf{49.51} & 48.04 & \textbf{60.76} & 60.39 \\
MedXpertQA & \textbf{11.88} & 11.59 & \textbf{19.31} & 18.45 \\
PubMedQA & \textbf{74.00} & 73.30 & 77.60 & \textbf{78.50} \\
\midrule
\textbf{Average} & \textbf{42.62} & 41.95 & 51.90 & \textbf{52.14} \\
\bottomrule
\end{tabular}
}
\caption{Ablation on the strength estimator. We fix the generator and compare different models for estimating distractor strength. ``Self'' indicates that the estimator is the same model as the one being trained (e.g., using Qwen2-7B to estimate for Qwen2-7B), while other names indicate cross-model estimation.}
\label{tab:estimator_ablation}
\end{table}

\section{Option-Label Distribution Analysis}
\label{sec:option_label_bias}

\begin{figure}[t]
    \centering
    \begin{subfigure}[t]{\columnwidth}
        \centering
        \includegraphics[width=\columnwidth]{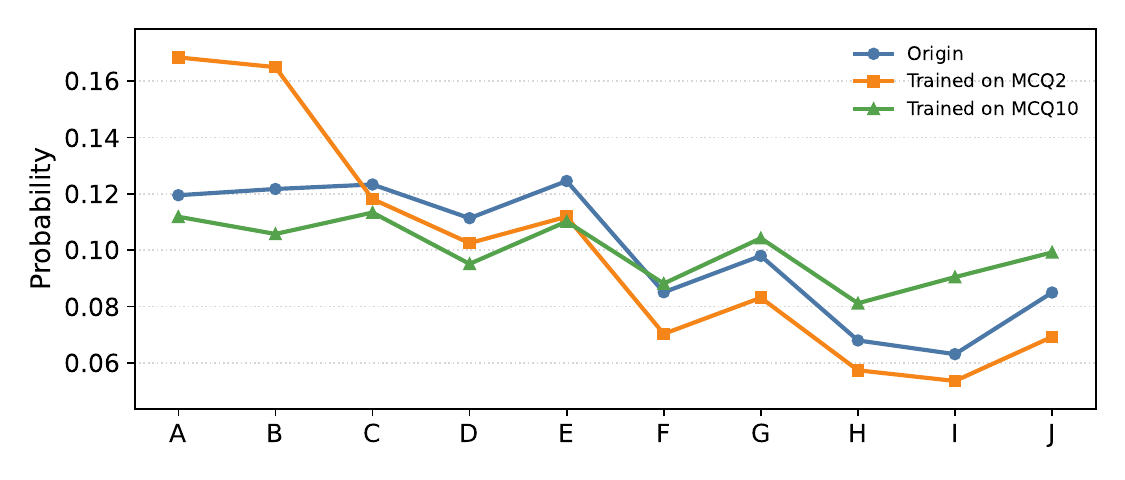}
        \subcaption{Qwen2-7B}
    \end{subfigure}
    \\
    \begin{subfigure}[t]{\columnwidth}
        \centering
        \includegraphics[width=\columnwidth]{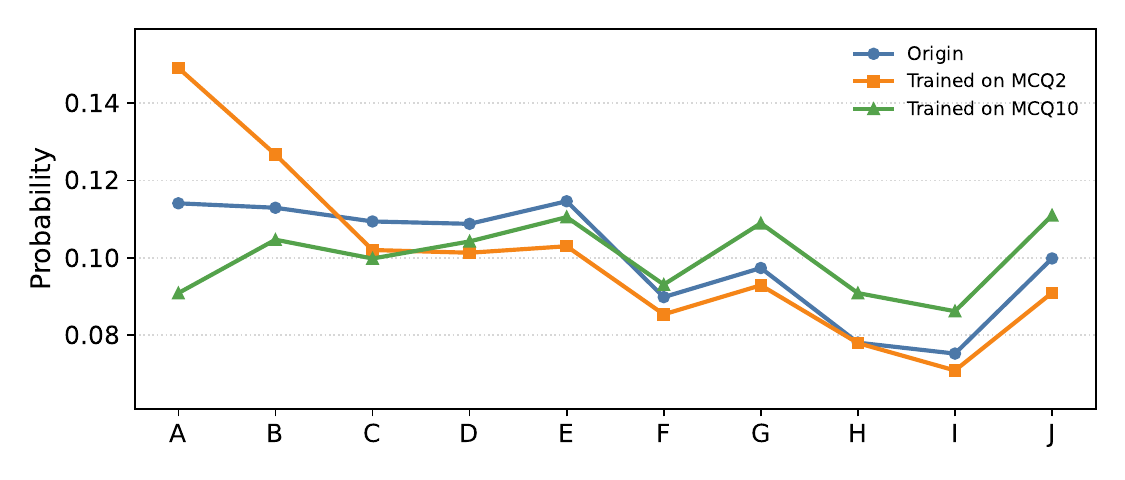}
        \subcaption{Llama-3.1-8B}
    \end{subfigure}
    \caption{Option-label distributions under origin vs. \texttt{mcq2} vs. \texttt{mcq10} training. Both models show A/B bias after \texttt{mcq2} training, a near-uniform distribution after \texttt{mcq10}, and origin in between.}
    \label{fig:line_option_label_distribution}
\end{figure}

\noindent\textbf{Experiment Setting.} We quantify option-label bias under controlled conditions.
Starting from the 10-choice test split of MMLU-Pro, we permute the correct label so that it is uniformly distributed over A--J.
For each item, we query the model with temperature 0.7 and draw 8 responses.
We aggregate the selected labels across the entire test split to obtain the empirical distribution.
Figure~\ref{fig:line_option_label_distribution} overlays these distributions for three training strategies: the original model (origin), training on \texttt{mcq2}, and training on \texttt{mcq10}.

\noindent\textbf{Findings.} The pattern is consistent across Qwen and Llama.
After \texttt{mcq2} training, both models assign disproportionate mass to early labels A/B, producing a front-loaded curve.
After \texttt{mcq10} training, the curve flattens and approaches the uniform baseline.
The origin models lie in between.
This indicates a distributional mismatch: models adapt their output-label distribution to the option space seen during training, and neither fewer nor more options is inherently superior.

\section{Multi-Seed Results}
\label{sec:multi_seed}

To verify the statistical significance of IDC's gains, we conducted additional experiments with 3 independent random seeds for both the standard RLVR baseline (Orig.) and our IDC method. Table~\ref{tab:multi_seed} reports mean $\pm$ std.

\begin{table}[t]
\centering
\small
\resizebox{\columnwidth}{!}{
\begin{tabular}{l|cc|cc}
\toprule
\multirow{2}{*}{\textbf{Dataset}} & \multicolumn{2}{c|}{\textbf{Qwen2-7B}} & \multicolumn{2}{c}{\textbf{Llama-3.1-8B}} \\
\cmidrule(lr){2-3} \cmidrule(lr){4-5}
& \textbf{Orig.} & \textbf{IDC} & \textbf{Orig.} & \textbf{IDC} \\
\midrule
MedQA & 47.16$\pm$0.29 & \textbf{49.72$\pm$0.44} & 59.01$\pm$0.38 & \textbf{66.02$\pm$0.47} \\
SuperGPQA & 25.53$\pm$0.40 & \textbf{27.85$\pm$0.38} & 33.24$\pm$0.36 & \textbf{35.65$\pm$0.43} \\
MMLU-Pro & 44.35$\pm$0.32 & \textbf{49.43$\pm$0.49} & 58.53$\pm$0.39 & \textbf{60.70$\pm$0.48} \\
MedXpertQA & 10.20$\pm$0.33 & \textbf{11.82$\pm$0.37} & 14.25$\pm$0.43 & \textbf{19.22$\pm$0.43} \\
PubMedQA & 69.39$\pm$0.40 & \textbf{73.91$\pm$0.42} & 77.92$\pm$0.44 & 77.58$\pm$0.48 \\
\midrule
\textbf{Average} & 39.33$\pm$0.23 & \textbf{42.55$\pm$0.24} & 48.59$\pm$0.25 & \textbf{51.83$\pm$0.24} \\
\bottomrule
\end{tabular}
}
\caption{Multi-seed results (mean $\pm$ std over 3 independent runs) for the standard RLVR baseline (Orig.) and IDC. IDC provides stable and consistent gains across runs.}
\label{tab:multi_seed}
\end{table}

The results confirm that IDC provides consistent gains over the baseline across multiple runs, with non-overlapping intervals on most benchmarks.

\section{Mixed-Option Training}
\label{sec:mixed_option}

In Section~\ref{sec:number_distractor}, we showed that option-count alignment is important. A natural question is: what happens when training on a mixture of option counts? We construct a mixed training set from the original 10-choice MMLU-Pro questions, where each question is randomly assigned a variant with an equal proportion (20\% each) of 2-, 4-, 6-, 8-, and 10-choice questions.

\begin{table*}[t]
\centering
\small
\begin{tabular}{ll|ccccc|c}
\toprule
\textbf{Model} & \textbf{Train} & \textbf{Test-2} & \textbf{Test-4} & \textbf{Test-6} & \textbf{Test-8} & \textbf{Test-10} & \textbf{Avg} \\
\midrule
\multirow{3}{*}{Llama-3.1-8B} & mix & 80.01 & 70.21 & 62.86 & 59.46 & 57.56 & 66.02 \\
& Best single (mcq6) & 79.57 & 69.58 & \textbf{64.07} & 59.21 & 57.87 & 66.06 \\
& Single-strategy Avg & 78.71 & 69.36 & 63.12 & 58.40 & 56.66 & 65.25 \\
\midrule
\multirow{3}{*}{Qwen2-7B} & mix & 71.29 & 64.57 & 56.99 & 51.94 & 49.01 & 58.76 \\
& Best single (mcq4) & \textbf{72.15} & \textbf{65.15} & \textbf{58.22} & 51.92 & 48.57 & \textbf{59.20} \\
& Single-strategy Avg & 71.46 & 63.79 & 57.24 & 51.95 & 48.67 & 58.62 \\
\bottomrule
\end{tabular}
\caption{Mixed-option training results on MMLU-Pro. ``Single-strategy Avg'' denotes the average over five single-option-count training runs (mcq2, mcq4, mcq6, mcq8, mcq10). The mix strategy outperforms the single-strategy average and provides robust generalization without prior knowledge of the test distribution.}
\label{tab:mixed_option}
\end{table*}

On both models, the mix strategy outperforms the single-strategy average. On Llama, mix is essentially on par with the best single strategy (mcq6). The mix strategy provides robust generalization without requiring prior knowledge of the test distribution, making it a practical default when the test format is unknown or heterogeneous.

\section{Generalization to the Law Domain}
\label{sec:law_domain}

To validate that IDC generalizes beyond the medical domain, we conducted an experiment on the law subset of MMLU-Pro ($\sim$1k samples, 85:5:10 train/dev/test split) using Qwen2.5-14B-Instruct. Due to the relatively small test set, we report results averaged over 3 independent runs (Table~\ref{tab:law_domain}).

\begin{table}[t]
\centering
\small
\begin{tabular}{l|c}
\toprule
\textbf{Setting} & \textbf{Accuracy (mean $\pm$ std)} \\
\midrule
Base (Qwen2.5-14B-Instruct) & 61.72 $\pm$ 1.51\% \\
Orig. (standard RLVR) & 68.32 $\pm$ 0.57\% \\
IDC & \textbf{71.05 $\pm$ 1.14\%} \\
\bottomrule
\end{tabular}
\caption{IDC results on the law subset of MMLU-Pro using Qwen2.5-14B-Instruct (3-seed average).}
\label{tab:law_domain}
\end{table}

IDC yields a clear improvement over standard RLVR in the law domain (+2.73\%), confirming that its benefits are not restricted to the medical domain.

\section{Option-Count Mismatch on Larger Models}
\label{sec:mismatch_14b}

To address whether option-count mismatch persists for larger models, we conducted additional experiments with Qwen2.5-14B-Instruct on MMLU-Pro (Table~\ref{tab:mismatch_14b}).

\begin{table}[t]
\centering
\small
\begin{tabular}{l|cc}
\toprule
\textbf{Model} & \textbf{mcq2-test} & \textbf{mcq10-test} \\
\midrule
Qwen2.5-14B (base) & 81.88 & 62.63 \\
\quad + mcq2-train & \textbf{86.13} & 62.92 \\
\quad + mcq10-train & 83.79 & \textbf{67.94} \\
\bottomrule
\end{tabular}
\caption{Option-count mismatch persists in Qwen2.5-14B-Instruct. The mcq2-trained model excels on mcq2-test but shows minimal improvement on mcq10-test, and vice versa.}
\label{tab:mismatch_14b}
\end{table}

The mismatch pattern is consistent with our findings on smaller models: the mcq2-trained model excels on mcq2-test but barely improves on mcq10-test, while the mcq10-trained model shows the reverse pattern. This confirms that option-count mismatch is not mitigated by scaling up model size, as it stems from a fundamental property of LLMs rather than insufficient model capacity.

\section{Algorithm Details}
\label{sec:algorithm_details}

We provide the detailed algorithm for Iterative Distractor Curation in Algorithm~\ref{alg:progressive-accumulation}.

\begin{algorithm*}[t]
\caption{Iterative Distractor Curation (IDC)}
\label{alg:progressive-accumulation}

\textbf{Input:} Question stem $q$; Correct answer $o^\ast$; Original distractors $\mathcal{D}_{orig}$; \\
\hspace*{1.65em} Generator model $\mathcal{M}_{gen}$; Evaluator model $\mathcal{M}_{eval}$; \\
\hspace*{1.65em} Target option count $N = |\mathcal{D}_{orig}| + 1$; Max iterations $T$; \\

\textbf{Output:} Optimized distractor set $\mathcal{D}_{final}$.

\begin{enumerate}
    \item \textbf{Initialization:} 
    \begin{enumerate}
        \item Construct initial option set $\mathcal{C}_0 = \{o^\ast\} \cup \mathcal{D}_{orig}$.
        \item Evaluate $\mathcal{C}_0$ using $\mathcal{M}_{eval}$ to obtain empirical strengths $\hat{s}(\cdot)$. Specifically, for each distractor $d \in \mathcal{D}_{orig}$, sample $K$ trajectories and compute:
        \[
        N_d = \sum_{k=1}^{K} \mathbb{I}[a_k = d,\, a_k \neq o^*], \quad
        N_{\text{err}} = \sum_{k=1}^{K} \mathbb{I}[a_k \neq o^*],
        \]
        then define $\hat{s}(d) = N_d / N_{\text{err}}$ if $N_{\text{err}} > 0$, otherwise $\hat{s}(d) = 0$.
        \item Initialize effective pool $\mathcal{D} \leftarrow \{d \in \mathcal{D}_{orig} \mid \hat{s}(d) > 0\}$.
        \item Initialize history $\mathcal{H} \leftarrow \{ (\mathcal{D}, p_0) \}$.
    \end{enumerate}

    \item \textbf{Iterative Optimization:} For $t = 1$ to $T$:
    \begin{enumerate}
        \item Let target distractor count $K_d = N - 1$.
        \item \textbf{Mode 1: Filling} (If $|\mathcal{D}| < K_d$):
        \begin{enumerate}
            \item Calculate needed count $k = K_d - |\mathcal{D}|$.
            \item Generate $k$ candidates $\mathcal{S}_{new}$ using $\mathcal{M}_{gen}(q, o^\ast, \text{exclude}=\mathcal{D})$.
            \item Construct test set $\mathcal{C}_t = \{o^\ast\} \cup \mathcal{D} \cup \mathcal{S}_{new}$.
            \item Evaluate $\mathcal{C}_t$ with $\mathcal{M}_{eval}$ to obtain strengths $\hat{s}_t(\cdot)$.
            \item Identify new effective items: $\mathcal{D} \leftarrow \mathcal{D} \cup \{d \in \mathcal{S}_{new} \mid \hat{s}_t(d) > 0\}$.
        \end{enumerate}
        
        \item \textbf{Mode 2: Replacement} (If $|\mathcal{D}| \ge K_d$):
        \begin{enumerate}
            \item Identify weakest link $d_{weak} = \arg\min_{d \in \mathcal{D}} \hat{s}(d)$.
            \item Generate 1 candidate $d_{new}$ using $\mathcal{M}_{gen}$.
            \item Construct test set $\mathcal{C}_t = \{o^\ast\} \cup (\mathcal{D} \setminus \{d_{weak}\}) \cup \{d_{new}\}$.
            \item Evaluate $\mathcal{C}_t$ with $\mathcal{M}_{eval}$ to obtain strengths $\hat{s}_t(\cdot)$.
            \item If $\hat{s}_t(d_{new}) > \hat{s}(d_{weak})$, then update pool: \\
                  $\mathcal{D} \leftarrow (\mathcal{D} \setminus \{d_{weak}\}) \cup \{d_{new}\}$.
        \end{enumerate}
        \item Record current state to history: $\mathcal{H} \leftarrow \mathcal{H} \cup \{ (\mathcal{D}, p_t) \}$.
    \end{enumerate}

    \item \textbf{Selection:}
    \begin{enumerate}
        \item Find maximum pool size $S_{max} = \max_{(\mathcal{D}, \cdot) \in \mathcal{H}} |\mathcal{D}|$.
        \item Select version $(\mathcal{D}^\ast, p^\ast) \in \mathcal{H}$ where $|\mathcal{D}^\ast| = S_{max}$ and $p^\ast$ is minimized.
        \item Set $\mathcal{D}_{final} \leftarrow \mathcal{D}^\ast$.
        \item While $|\mathcal{D}_{final}| < K_d$, fill with unreplaced options from $\mathcal{D}_{orig}$.
    \end{enumerate}

    \item Return $\mathcal{D}_{final}$.

\end{enumerate}
\end{algorithm*}

\end{CJK}  % 结束中文环境
\end{document}